%% file: holo360.tex
\documentclass[10pt,twocolumn,letterpaper]{article}

\usepackage[pagenumbers]{cvpr}      

\usepackage{booktabs}
\usepackage{multirow}
\usepackage{array}
\usepackage{amsmath}
\usepackage{adjustbox}
\usepackage{amssymb} 
\usepackage{wrapfig}
\usepackage{pifont}   
\usepackage{xcolor}   
\usepackage{caption}

\input{preamble}
\definecolor{cvprblue}{rgb}{0.21,0.49,0.74}
\usepackage[pagebackref,breaklinks,colorlinks,allcolors=cvprblue]{hyperref}

\def\paperID{*****} 
\def\confName{CVPR}
\def\confYear{2026}

\title{Holo360D: A Large-Scale Real-World Dataset with Continuous Trajectories for Advancing Panoramic 3D Reconstruction and Beyond}

\author{
Jing OU$^{1,*}$ \quad Zidong Cao$^{1,*}$ \quad Yinrui Ren$^{1,2}$ \quad Zhuoxiao Li$^{1}$ \quad Jinjing Zhu$^{1}$ \\
Tongyan Hua$^{1}$ \quad Shuai Zhang$^{1}$ \quad Hui Xiong$^{1,\dagger}$ \quad Wufan Zhao$^{1,\dagger}$ \\
$^{1}$The Hong Kong University of Science and Technology (Guangzhou) \\
$^{2}$South China Normal University \\
$^*$Equal Contribution \quad $^\dagger$Corresponding Author
}

\begin{document}

\makeatletter
\apptocmd{\@maketitle}{
  \par\noindent
  \begin{minipage}{\textwidth}
    \centering
    \vspace{5pt}
    \includegraphics[width=0.99\textwidth]{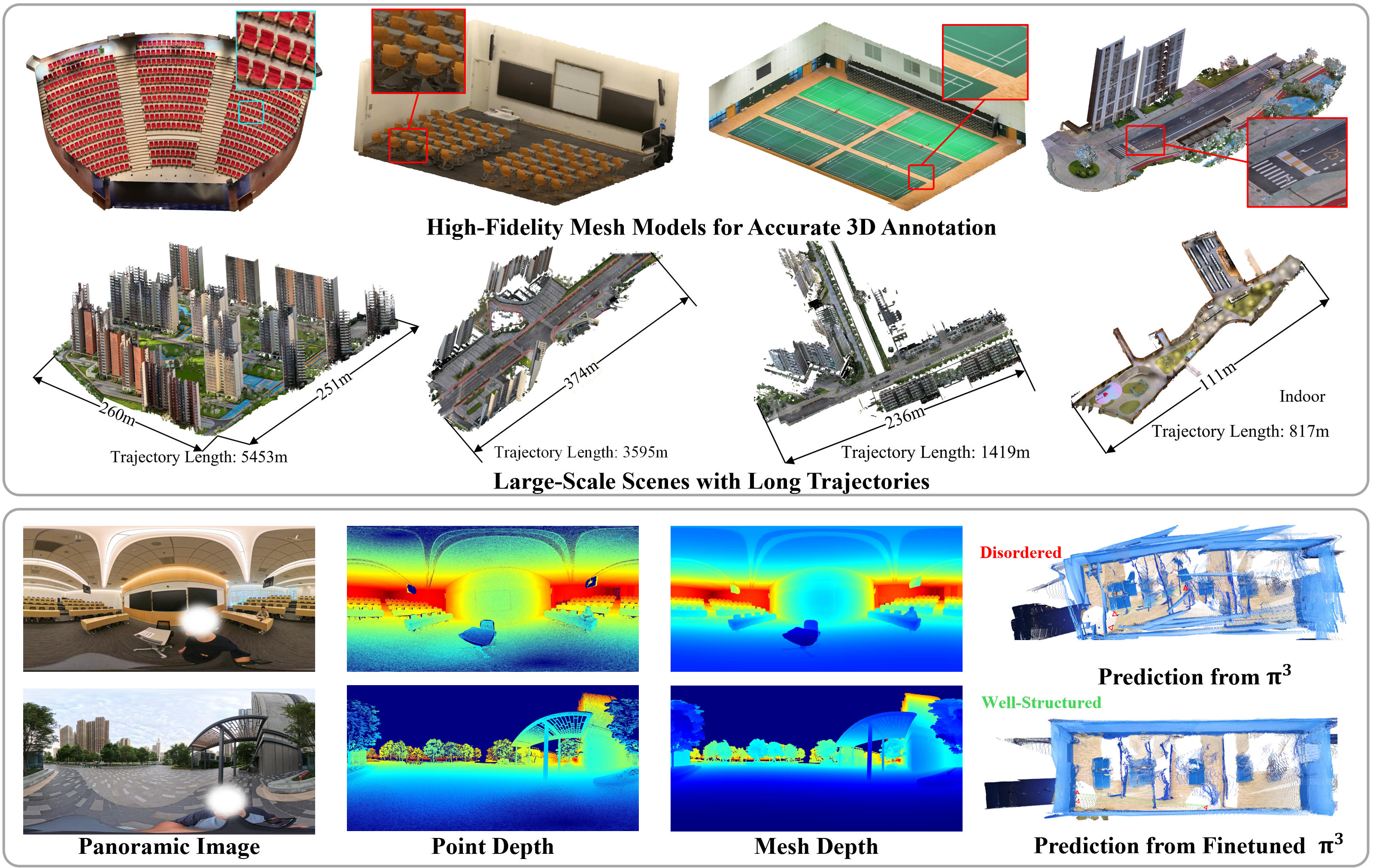}
    \captionof{figure}{We present \textbf{Holo360D}, the first large-scale real-world panoramic 3D dataset, containing 109,495 panoramas paired with LiDAR-derived ground truth, including precise meshes, point clouds, depth maps, and camera poses. More importantly, Holo360D is the first panoramic dataset to offer \textbf{accurately aligned high-completeness} depth maps with \textbf{continuous} camera trajectories over long sequences.}
    \label{fig:coverfigure}
    \vspace{10pt}
  \end{minipage}
  \par
}{}{}
\makeatother

\maketitle
\input{0_abstract}
\input{1_intro}

\input{2_relatedworks}
\input{3_theHolo360Dataset}

\input{4_experiment}

\input{5_conclusions}
{
    \small
    \bibliographystyle{ieeenat_fullname}
    \bibliography{main}
}
\input{X_suppl}

\end{document}

%% file: 0_abstract.tex
\begin{abstract}
While feed-forward 3D reconstruction models have advanced rapidly, they still exhibit degraded performance on panoramas due to spherical distortions. Moreover, existing panoramic 3D datasets are predominantly collected with $360^{\circ}$ cameras fixed at discrete locations, resulting in \textit{\textbf{discontinuous trajectories}}. These limitations critically hinder the development of panoramic feed-forward 3D reconstruction, especially for the multi-view setting. In this paper, we present \textbf{Holo360D}, a comprehensive dataset containing 109,495 panoramas paired with registered point clouds, meshes, and aligned camera poses. To our knowledge, Holo360D is the first large-scale dataset that provides \textit{\textbf{continuous panoramic sequences}} with \textit{\textbf{accurately aligned high-completeness depth maps}}. The raw data are initially collected using a 3D laser scanner coupled with a $360^{\circ}$ camera. 
Subsequently, the raw data are processed with both online and offline SLAM systems. Furthermore, to enhance the 3D data quality, a post-processing pipeline tailored for the 360$^\circ$ dataset is proposed, including geometry denoising, mesh hole filling, and region-specific remeshing, \textit{etc}. Finally, we establish a new benchmark by fine-tuning 3D reconstruction models on Holo360D, providing key insights into effective fine-tuning strategies. Our results demonstrate that Holo360D delivers superior training signals and provides a comprehensive benchmark for advancing panoramic 3D reconstruction models. Datasets and Code will be made publicly available. Github page: \url{https://github.com/Jou719/Holo360D}.
\end{abstract}

%% file: 1_intro.tex
\section{Introduction}
\label{sec:intro}

In recent years, feed-forward 3D reconstruction models~\cite{wang2024dust3r, wang2025pi, yang2025fast3r, zhang2024monst3r, wang2025vggt} have advanced significantly by scaling up both model capacity and training data, yielding superior performance on tasks such as monocular depth estimation~\cite{wang2025moge,wang2024depth} and multi-view reconstruction~\cite{yang2025fast3r,leroy2024grounding,wang2025vggt,wang2025pi}. For example, VGGT~\cite{wang2025vggt}, a feed-forward transformer-based architecture, can jointly predict camera poses, depth maps, and point maps from multi-view perspective images. However, most existing 3D models are developed for perspective images, and their performance degrades significantly when applied to panoramic images~\cite{cao2024panda, wang2024depth}. The degradation arises primarily from the spherical distortion inherent in panoramas, where the widely adopted equirectangular projection introduces non-uniform sampling and severe stretching near poles~\cite{ai2025survey,cao2025st}. Thus, the geometric priors learned from perspective images are no longer valid under such distortions.

Existing panoramic 3D datasets~\cite{armeni2017joint,zioulis2018omnidepth,chang2017matterport3d,zheng2020structured3d,li2022mode,huang2024360loc, albanis2021pano3d} suffer from substantial limitations in scale, depth map quality, and viewpoint continuity. \textbf{(I) Scale:} Popular datasets such as Stanford2D3D~\cite{armeni2017joint} and Matterport3D~\cite{chang2017matterport3d} contain fewer than 11K panoramic samples, making effective fine-tuning challenging. \textbf{(II) Depth Map Quality:} Depth maps in existing datasets such as Matterport3D~\cite{chang2017matterport3d} and KITTI-360~\cite{liao2022kitti} are often incomplete due to insufficient scans, including occluded areas and glass regions. The alignment accuracy between depth maps and RGB images is also limited, especially under outdoor scenes~\cite{huang2024360loc,liao2022kitti}. \textbf{(III) Trajectory Continuity:} The panoramic images in Matterport3D are stitched from perspective views captured at fixed and discrete locations. These locations are spaced 2.25m apart on average~\cite{chang2017matterport3d}, resulting in a sparse and wide-baseline setting. Consequently, existing multi-view methods~\cite{chen2023panogrf,zhang2025pansplat,chen2025splatter} designed for panoramas are typically restricted to configurations with very few input views, \eg, two views.
\begin{table*}[t]
\footnotesize
    \setlength\tabcolsep{2pt}
    \centering 
    \renewcommand\arraystretch{1.2}
    
\centering
\caption{Comparison of panoramic 3D datasets. Holo360D is the only large-scale real-world panoramic dataset that provides accurately aligned high-completeness depth maps and continuous camera trajectories. Continuity: availability of continuous panoramic sequences (Average inter-frame distance to quantify continuity). Alignment: depth–panorama alignment error. Depth Completion: proportion of valid depth pixels in the depth map. "I" and "O": Indoor and Outdoor. \textit{All metrics are computed as described in ~\cref{sec:Analysis of the Dataset}. We do not report some metrics for Depth360 because we could not obtain the data despite multiple download requests.}}
\label{tab:summary_datasets}
\begin{adjustbox}{max width=\linewidth}
  \input{summarydataset}
\end{adjustbox}
\end{table*}
To address these challenges, we introduce \textit{Holo360D}, a large-scale real-world panoramic 3D dataset with 109,495 panoramas, featuring continuous camera trajectories and accurately aligned high-completeness depth maps (see ~\cref{tab:summary_datasets} and ~\cref{fig:depth comparison}). The data are captured using a handheld laser scanner coupled with a 360$^{\circ}$ camera. The scanner integrates LiDAR, three pinhole cameras, IMU, and RTK-GNSS for precise localization in outdoor scenarios. The 360$^{\circ}$ camera is rigidly mounted on top of the scanner, and both the scanner and camera share a unified software trigger to ensure synchronous data recording. Although LiDAR excels at long-range sensing, its resulting point clouds are often sparse under high-speed motion and limited viewpoint coverage. To mitigate this sparsity, the data capture process maintains a gradual motion (about 0.3 m/s indoors and 0.6 m/s outdoors). In addition, we employ a continuous traversal strategy with overlapping trajectories for each scene to enhance point cloud completeness. As a result, the data recording spans over 19 hours, with a total trajectory distance exceeding 31 $\text{km}$. The outdoor scenes cover an area greater than 0.17 $\text{km}^2$. The raw data include panoramic images, point clouds, and camera poses.

We then process the raw data in several stages. First, we utilize an onboard SLAM system embedded in the 3D laser scanner, which processes the raw data to generate coarsely registered point clouds in real-time. Subsequently, we feed the coarsely registered point clouds, along with camera poses and IMU measurements, using a high-precision offline SLAM system to jointly generate aligned camera poses and registered point clouds. In addition, we perform surface reconstruction on the registered point clouds for each scene to produce a dense and consistent mesh model.

Despite the careful acquisition and registration process, the initial output meshes still contain artifacts, such as outliers and incomplete regions. To address these artifacts, we design a data post-processing pipeline consisting of three steps: (i) data denoising to remove isolated points, (ii) mesh completion to fill in glass and occluded regions, and (iii) region-specific remeshing to better preserve thin structures. 

Finally, we establish a new benchmark by fine-tuning leading feed-forward 3D reconstruction models~\cite{zhang2025flare, wang2025vggt, wang2025pi} on Holo360D. We empirically identify effective fine-tuning strategies, such as joint supervision from point clouds and meshes. Experimental results demonstrate that Holo360D provides superior training signals compared to previous panoramic 3D datasets~\cite{chang2017matterport3d}. More importantly, our Holo360D dataset offers a comprehensive benchmark for advancing panoramic feed-forward 3D reconstruction models in both training and evaluation.

The main contributions of our work are summarized as follows: 
\begin{itemize}
\item We propose \textit{Holo360D}, a large-scale real-world panoramic 3D dataset with continuous trajectories and accurately aligned high-completeness depth maps; 

\item We propose a data post-processing pipeline, including data denoising, mesh hole filling, and region-specific remeshing, which produces high-quality depth maps; 

\item We establish a new benchmark by fine-tuning leading feed-forward 3D reconstruction models on Holo360D. The results demonstrate that Holo360D provides superior training signals.
\end{itemize}

%% file: summarydataset.tex
\begin{tabular}{l|c|c|c|c|c|c|c}
\toprule
\textbf{Datasets}  & Indoor & Outdoor & Continuity & Alignment$\downarrow$ & Depth Completion$\uparrow$ & Scenes & Numbers \\
\midrule
Stanford2D3D~\cite{armeni2017joint}  
& \textcolor{green}{\ding{51}} & \textcolor{red}{\ding{55}} & \textcolor{red}{\ding{55}} & 9.45 & 0.72(I) & 10 & 1,314 \\

Matterport3D~\cite{chang2017matterport3d}  
& \textcolor{green}{\ding{51}} & \textcolor{red}{\ding{55}} & \textcolor{red}{\ding{55}} & 7.99 & 0.62(I) & \textbf{90} & 10,790 \\


Depth360~\cite{Feng2022360depth360}   
& \textcolor{red}{\ding{55}} & \textcolor{green}{\ding{51}} & \textcolor{green}{\ding{51}}(N/A) & N/A & N/A & 30 & 30,000 \\
360Loc~\cite{huang2024360loc} 
& \textcolor{green}{\ding{51}} & \textcolor{green}{\ding{51}} & \textcolor{green}{\ding{51}}(0.49) & 12.24 & 0.62(I), 0.7(O) & 4 & 2,244 \\

KITTI-360~\cite{liao2022kitti}   
& \textcolor{red}{\ding{55}} & \textcolor{green}{\ding{51}} & \textcolor{green}{\ding{51}}(1.01) & 11.72 &  0.16(O) & 11 & 83,000 \\

\midrule
\textbf{Holo360D(Ours)} 
& \textcolor{green}{\ding{51}} & \textcolor{green}{\ding{51}} & \textcolor{green}{\ding{51}}(\textbf{0.29}) & \textbf{5.03} & \textbf{0.86(I), 0.82(O)} & 75 & \textbf{109,495} \\

\bottomrule
\end{tabular}

%% file: 2_relatedworks.tex
\section{Related Works}
\label{sec:related_works}

\begin{figure}[t]
    \centering
    \includegraphics[width=1\linewidth]{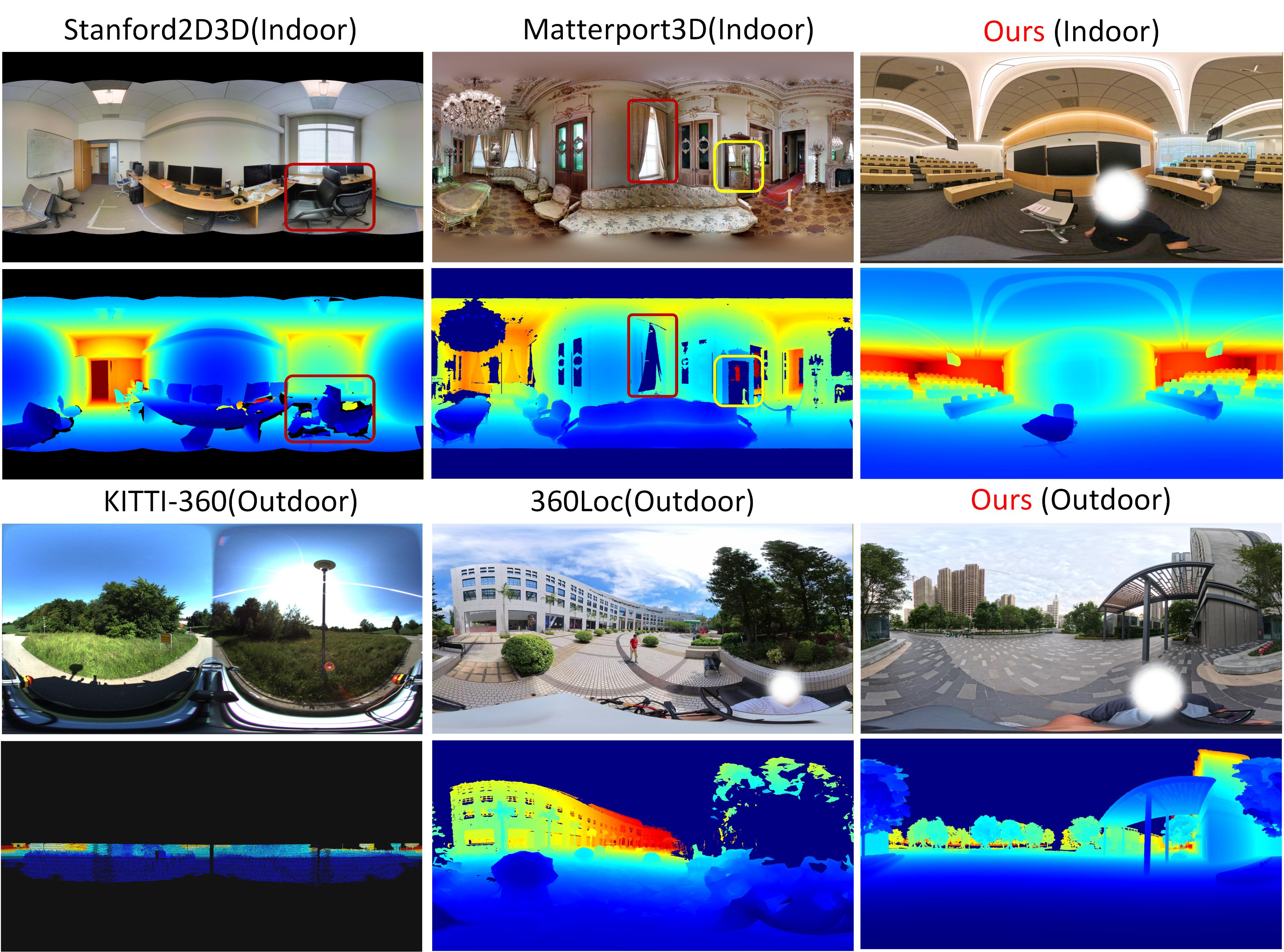}
    \caption{Comparison of depth maps across different panoramic datasets. Holo360D provides the highest-quality depth maps for both indoor and outdoor environments.}
    \label{fig:depth comparison}
\end{figure}

\subsection{Panoramic 3D Datasets}
\label{subsec:panoramic-3d-dataset}

Recent advances in large-scale and high-quality panoramic 3D datasets have stimulated the emergence of 3D reconstruction models~\cite{jung2025im360, chang2017matterport3d, huang2024360loc, li2022mode,armeni2017joint,zheng2020structured3d}. These models demonstrate remarkable zero-shot generalization capabilities through training on diverse scenarios. However, existing panoramic 3D datasets exhibit three main limitations. \textbf{(I) Scale Constraints:} Widely-used indoor datasets like Matterport3D~\cite{chang2017matterport3d} contain fewer than 11K panoramic samples, while outdoor datasets such as Deep360~\cite{li2022mode} typically comprise several thousand samples. This limited scale significantly hinders effective fine-tuning of feed-forward 3D reconstruction models. Although synthetic datasets can generate large quantities of samples, models trained on synthetic data often fail to generalize to real-world scenarios due to domain gaps. \textbf{(II) Limited Depth Quality:} The depth maps provided by existing panoramic 3D datasets often exhibit limited completeness~\cite{chang2017matterport3d,liao2022kitti} and suboptimal alignment accuracy~\cite{huang2024360loc, liao2022kitti}. This is mainly due to (i) incomplete geometry (meshes or point clouds) used to render depth, which introduces missing depth values, and (ii) insufficient camera pose accuracy, which leads to depth–panorama misalignment;  \textbf{(III) Trajectory Discontinuity:} Existing datasets typically consist of discrete panoramic captures from fixed locations, inherently limiting multi-view methods to wide-baseline settings with minimal input views (\eg, 2-3 views). While 360Loc~\cite{huang2024360loc} introduces continuous viewpoint trajectories, its primary focus on visual localization results in suboptimal depth quality without dedicated refinement. Another approach~\cite{Feng2022360depth360} employing Structure-from-Motion (SfM)~\cite{cui2017hsfm,hartley2003multiple} on 360$^{\circ}$ video sequences produces continuous trajectories, but the reconstructed depth maps remain inferior to those acquired with professional 3D scanning equipment.\textit{ To address these challenges, we propose a large-scale panoramic 3D dataset that provides continuous panoramic sequences with accurately aligned high-completeness depth maps.}
\begin{figure*}[t]
    \centering
    \includegraphics[width=1\linewidth]{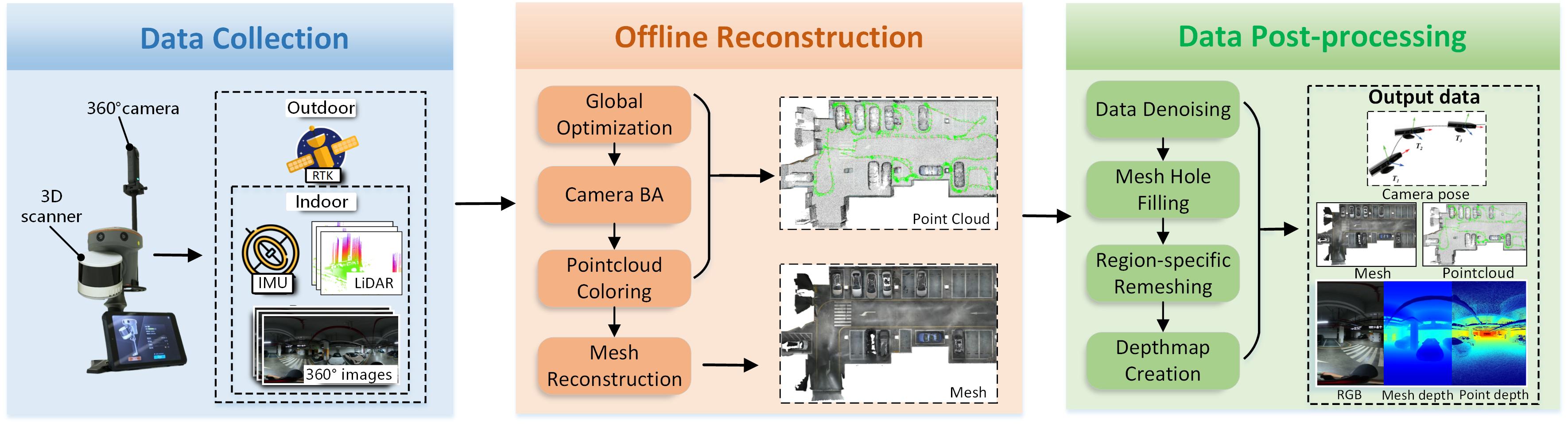}
    \caption{Dataset creation pipeline consisting of (i) data collection, (ii) offline reconstruction, and (iii) data post-processing.}
    \label{fig:datacapture}
\end{figure*}

\subsection{Feed-forward 3D Reconstruction Models}
\label{subsec:panoramic-3d-models}

DUST3R~\cite{wang2024dust3r} pioneers a novel paradigm for geometric understanding by demonstrating that multi-dataset training enables direct recovery of 3D properties (\textit{e.g.}, point maps and camera poses) from uncalibrated multi-view perspective images. 
%
%
Following DUST3R's groundbreaking paradigm shift, several works have extended its paradigm for enhanced geometric understanding. MASt3R~\cite{leroy2024grounding} augments DUST3R with a dense local feature extraction head, improving robustness in image matching. Spann3R~\cite{wang20243d} introduces a spatial memory network to handle multi-view inputs efficiently, eliminating the need for global alignment. Fast3R~\cite{yang2025fast3r} overcomes sequential limitations by processing multiple views simultaneously via a global fusion transformer, significantly boosting reconstruction quality. CUT3R~\cite{wang2025continuous} maintains a persistent scene state for incremental updates, supporting both static and dynamic scenes. SLAM3R~\cite{liu2025slam3r} enables real-time dense reconstruction from monocular videos by extending DUST3R to multi-view inputs. 
%
VGGT~\cite{wang2025vggt} scales the approach further with a large transformer that jointly predicts point clouds, camera poses, and intrinsics in a single forward pass.
Trained on millions of 3D samples spanning diverse environments~\cite{reizenstein2021common,pan2023aria,ling2024dl3dv,li2018megadepth,antequera2020mapillary,huang2018deepmvs}, VGGT achieves state-of-the-art performance in wild settings. However, its architecture imposes a critical dependency on the reference frame. The performance degrades significantly when provided with low-quality reference views. To address this fundamental limitation, ${\pi}^3$~\cite{wang2025pi} introduces a permutation-equivariant architecture that eliminates reference frame bias. 
 However, a key challenge for ${\pi}^3$ and similar emerging models lies in their lack of fine-tuning on domain-specific datasets. This limitation is acutely evident in panoramic 3D reconstruction, where the field remains predominantly constrained to wide-baseline configurations primarily due to the scarcity of training data featuring continuous viewpoint trajectories, hindering the application of the general models to such specialized tasks.

%% file: 3_theHolo360Dataset.tex
\section{The Holo360D Dataset}
\label{sec:dataset}

We introduce a large-scale panoramic 3D dataset featuring continuous panoramic sequences with high-quality depth maps. As illustrated in ~\cref{fig:datacapture}, the data collection process involves synchronized capture of panoramic images, point clouds and camera poses using a handheld platform (See~\cref{sec:Data Acquisition}). Through offline reconstruction and post-processing, we further generate refined camera poses, high-quality meshes, and dense panoramic depth maps (See~\cref{sec:Offline reconstruction} and~\cref{sec:Data Post-processing}). Finally, we provide a detailed quantitative analysis of the data quality (See~\cref{sec:Analysis of the Dataset}).

\subsection{Data Collection }
\label{sec:Data Acquisition}
The Holo360D dataset is captured using a handheld system composed of a 3D laser scanner and a 360$^{\circ}$ camera, which are rigidly mounted and synchronized via a shared software trigger to simultaneously start and stop data recording. The scanner integrates LiDAR, RTK-GNSS, IMU, and pinhole cameras; its onboard SLAM system fuses multi-sensor data to produce coarsely aligned point clouds and camera poses. The 360$^{\circ}$ camera records high-resolution panoramic videos at 24 fps with a resolution of 5760$\times$2880.
\textit{Further technical specifications are provided in the Supplementary Material}.

\subsection{Offline Reconstruction}
\label{sec:Offline reconstruction}
With limited onboard compute resources and real-time constraints, the scanner outputs only coarse poses and point clouds. As illustrated in Stage 2 of ~\cref{fig:datacapture}, the data collected in ~\cref{sec:Data Acquisition} are fed into an offline reconstruction pipeline that begins with a global bundle adjustment, jointly refining 360$^{\circ}$ camera poses and precisely aligning point clouds. Next, panoramic images are projected to colorize the point cloud, recovering accurate appearance. Further, the Poisson surface reconstruction is then performed on the point cloud to produce a mesh. At this stage, we output globally aligned point clouds and meshes, together with the images from the pinhole and 360° cameras and their corresponding poses.

\subsection{Data Post-processing}
\label{sec:Data Post-processing}
Even with careful scanning, directly using the point clouds and meshes still encounters several issues: (i) isolated outliers, (ii) incomplete meshes in glass and occluded regions, and (iii) low-quality geometry in thin structures. We address these with a three-step post-processing pipeline: denoising, mesh hole filling, and region-specific remeshing. Using the refined high-quality geometry, we then render accurate multi-view depth maps (See Stage 3 of ~\cref{fig:datacapture}).
\begin{figure*}[t]
    \centering
    \includegraphics[width=0.9\linewidth]{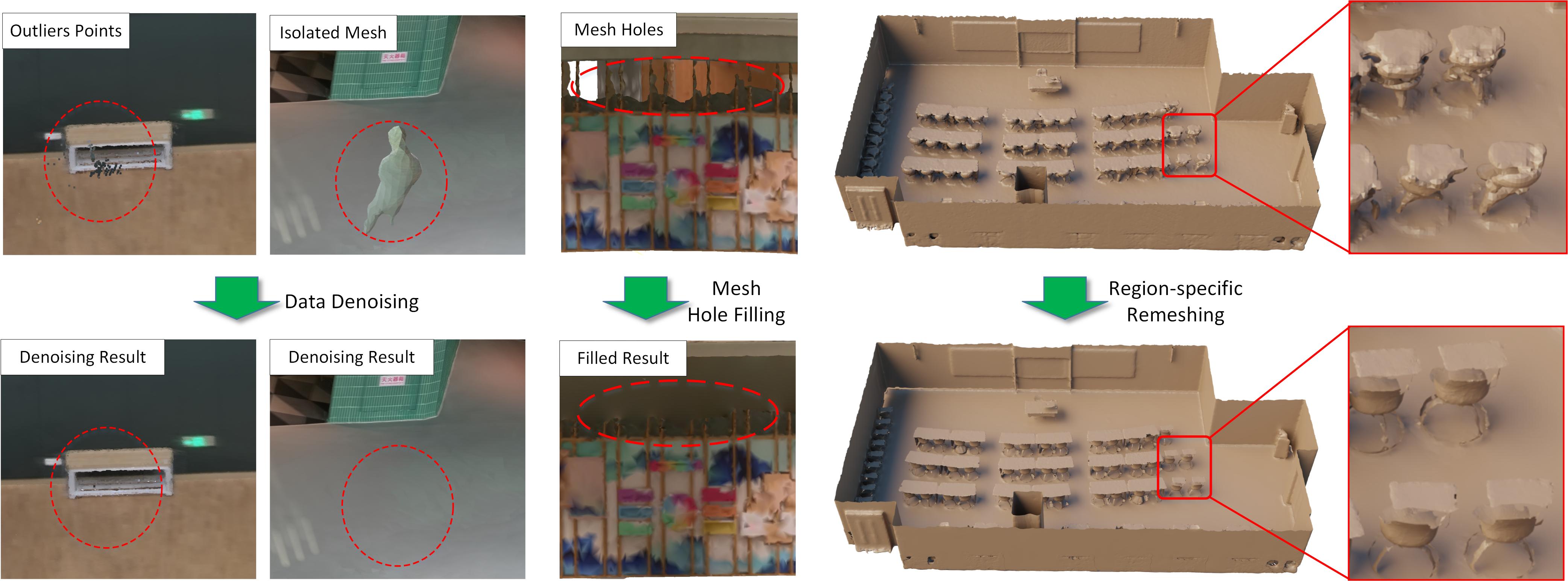}
    \caption{Data post-processing pipeline consisting of (i) data denoising, (ii) mesh hole filling, and (iii) region-specific remeshing.}
    \label{fig:Datapostprocessing}
\end{figure*}

\noindent\textbf{Data Denoising.} We collect data in residential areas to reflect real-world conditions, which in turn introduces two main error sources: (i) motion artifacts from dynamic pedestrians and (ii) specular reflection outliers from reflective surfaces. These artifacts propagate to the depth maps, producing incorrect depths.

We address these defects in the point clouds and meshes with a three-step denoising pipeline. First, we manually crop the point cloud to regions of interest to remove invalid data. Second, we apply radius outlier removal to eliminate isolated points. Finally, we visually inspect all scenes and remove any remaining large noise clusters and mesh patches. Denoising results are shown in ~\cref{fig:Datapostprocessing}.
\begin{figure*}[t]
    \centering
    \includegraphics[width=0.8\linewidth]{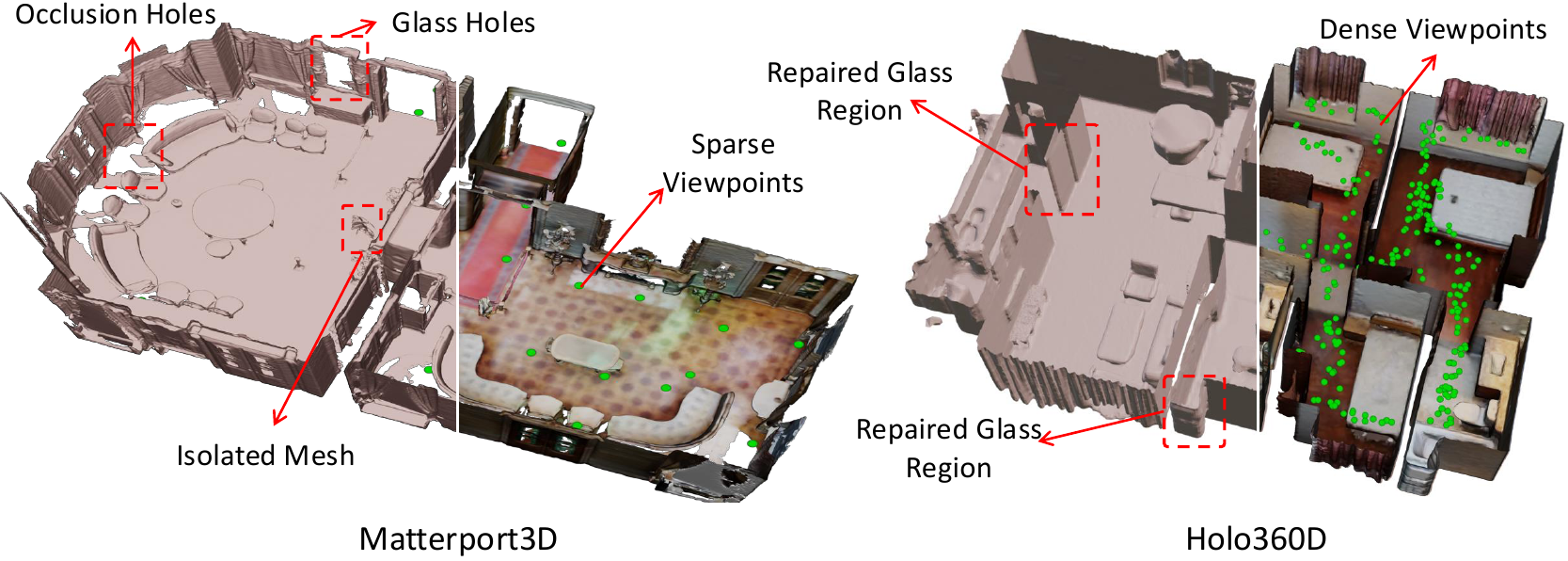}
    \caption{Comparison of reconstructed mesh models on Matterport3D and Holo360D. Holo360D meshes exhibit higher completeness in occluded and reflective glass regions and contain fewer floating artifacts.}
    \label{fig:mp3d-holo360d--mesh-pose}
\end{figure*}

\noindent\textbf{Mesh Hole Filling.} During the scanning of complex scenes, although our handheld system can reconstruct scenes from a wide range of viewpoints, occlusions are unavoidable, yielding incomplete meshes in occluded regions. In addition, transmission through glass, such as windows, leads to extremely sparse and unreliable point clouds. During mesh reconstruction, these sparse zones manifest as holes, which cause missing and erroneous depths in the resulting depth maps.

We complete the mesh via a three-step pipeline. First, we detect holes and measure the perimeter $P$ of each hole to quantify its size. Second, small holes are automatically filled using a curvature-preserving triangulation, ensuring that the resulting patches match the curvature of the surrounding mesh. Finally, for larger holes, we first insert bridge edges across the hole to subdivide it into smaller holes, and then apply curvature-preserving triangulation to each sub-hole to minimize geometric distortion. This strategy yields a model that is both complete and geometrically accurate, as shown in ~\cref{fig:Datapostprocessing}.

\noindent\textbf{Region-specific Remeshing.} During mesh reconstruction, point clouds are downsampled and smoothed to ensure computational efficiency and smoothness of surface. For thin-walled or complex objects, this removes critical details and degrades mesh quality, as shown in ~\cref{fig:Datapostprocessing}.

To address this issue, we implement a region-specific remeshing strategy. For regions with high reconstruction quality, such as walls and floors, we retain the mesh produced in ~\cref{sec:Data Post-processing}; for low-quality regions, such as furniture with complex structures, we remove these regions and reconstruct them using the original high-resolution point cloud. This strategy effectively controls overall computational cost while significantly enhancing geometric details and reconstruction completeness of the model.

\begin{table*}[t]
\small
    \setlength\tabcolsep{2pt}
    \centering 
    \renewcommand\arraystretch{1.2}
    \caption{Comparison of fine-tuning results under panorama and split views representations.}
     \resizebox{1.0\textwidth}{!}{
\begin{tabular}{lcccccc|cc|cccc}
\toprule
\multirow{2}{*}{} 
  & \multicolumn{3}{c}{Indoor} 
  & \multicolumn{3}{c|}{Outdoor} 
  & \multicolumn{1}{c}{Indoor} 
  & \multicolumn{1}{c|}{Outdoor}
  & \multicolumn{2}{c}{Indoor} 
  & \multicolumn{2}{c}{Outdoor} \\
\cmidrule(lr){2-4} \cmidrule(lr){5-7} \cmidrule(lr){8-8} \cmidrule(lr){9-9} \cmidrule(lr){10-11} \cmidrule(lr){12-13}
& ~~ATE$\downarrow$~~ & ~~RPE$_t$$\downarrow$~~ & ~~RPE$_r$$\downarrow$~~ 
& ~~ATE$\downarrow$~~ & ~~RPE$_t$$\downarrow$~~ & ~~RPE$_r$$\downarrow$~~ 
& \scriptsize AUC@30$\uparrow$ 
& \scriptsize AUC@30$\uparrow$ 
& ~~Acc.$\downarrow$~~ & ~~Comp.$\downarrow$~~ 
& ~~Acc.$\downarrow$~~ & ~~Comp.$\downarrow$~~ \\
\midrule
$\pi^3$ (baseline) & 0.093 & 0.112 & 2.034 & 0.201 & 0.228 & 0.960 & 0.733 & 0.758 & 0.052 & 0.033 & 0.293 & 0.221 \\
$\pi^3$ (panorama) & 0.018 & 0.030 & 0.621 & \textbf{0.064} & 0.110 & 0.260 & \textbf{0.961} & \textbf{0.945} & 0.075 & 0.160 & 0.718 & 0.584\\
$\pi^3$ (split views) & \textbf{0.014} & \textbf{0.014} & \textbf{0.295} & 0.065 & \textbf{0.049} & \textbf{0.208} & 0.837 & 0.818 & \textbf{0.024} & \textbf{0.018} & \textbf{0.194} & \textbf{0.152} \\
\bottomrule
\end{tabular}}
\label{tab:Input type}
\end{table*}

\begin{figure}[t]
    \centering
    \includegraphics[width=\columnwidth]{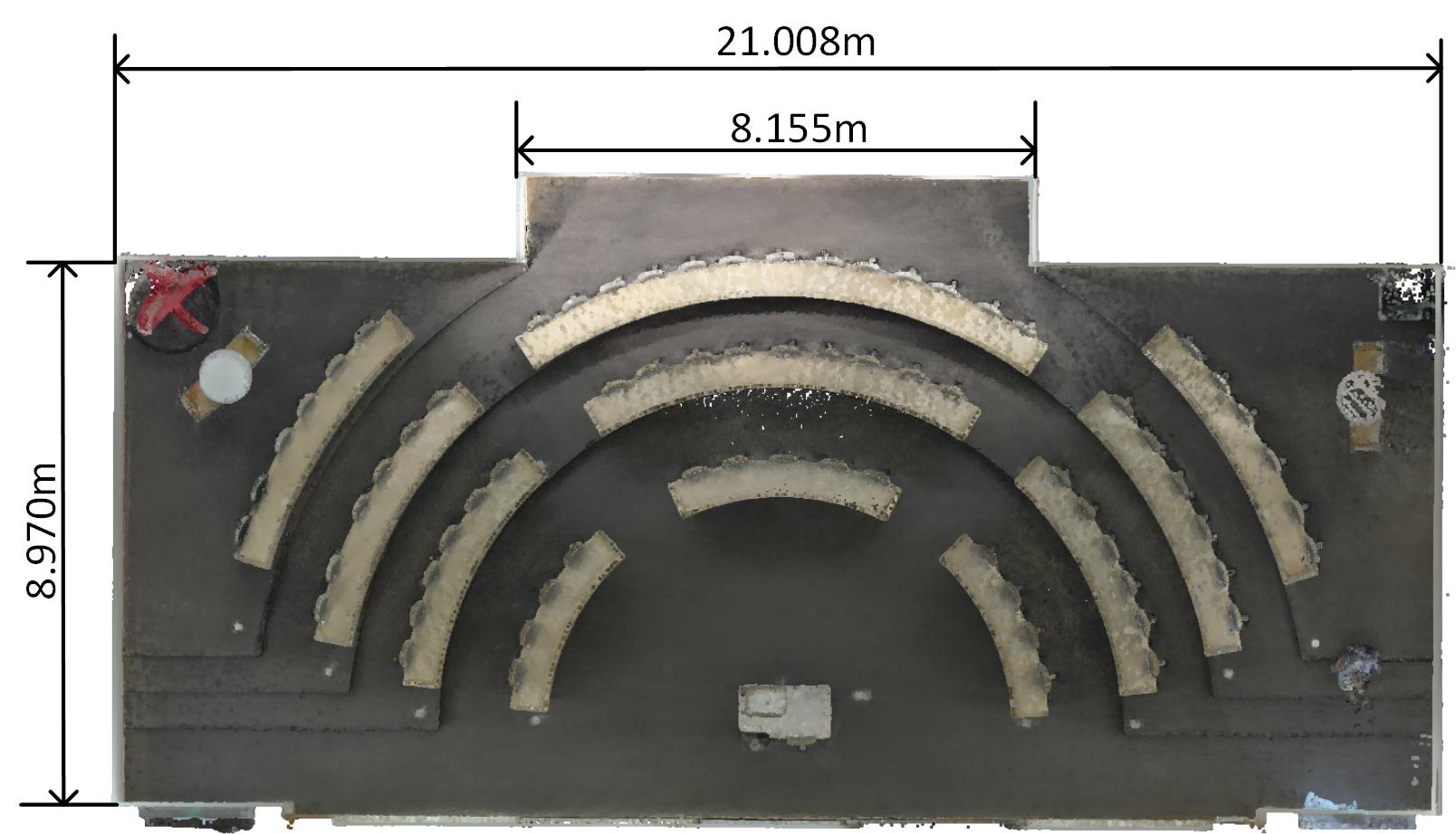}
    \caption{Reference dimensions used to evaluate point cloud reconstruction accuracy.}
    \label{fig:pointcloud_reconstruction_accuracy}
\end{figure}

\begin{figure}[t]
    \centering
    \includegraphics[width=\columnwidth]{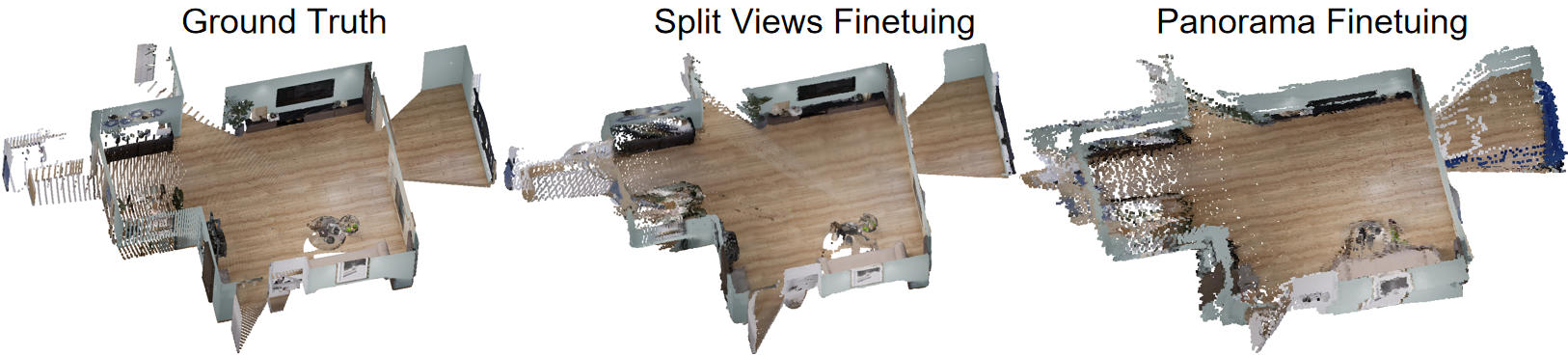}
    \caption{Visualization of fine-tuning performance with different input representations.}
    \label{fig:inputRepresenting}
\end{figure}
\begin{figure*}[t]
    \centering
    \includegraphics[width=0.98\linewidth]{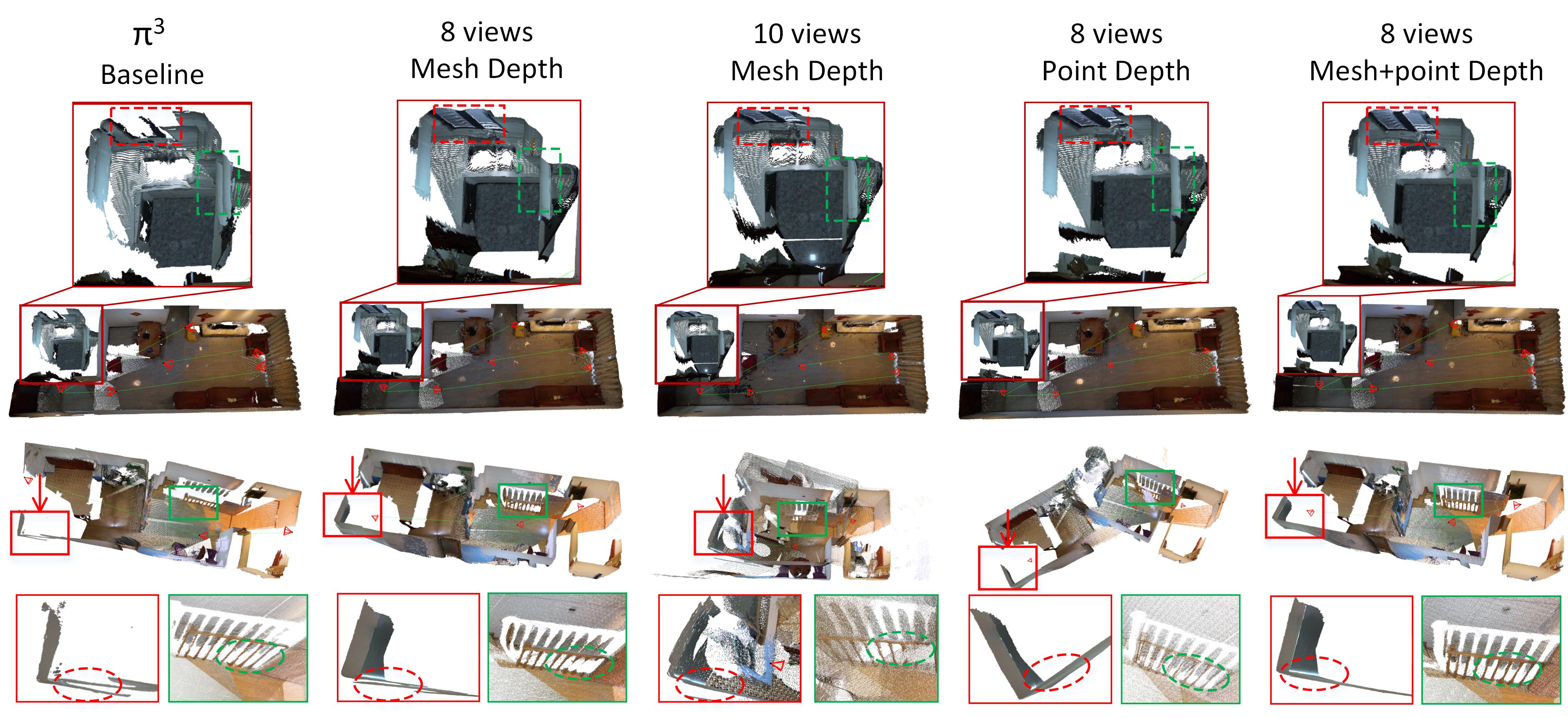}
    \caption{Visualization results comparing different view configurations and depth supervision types.}
    \label{fig:Ablatioion}
\end{figure*}
\begin{figure}[t]
    \centering
    \includegraphics[width=\linewidth]{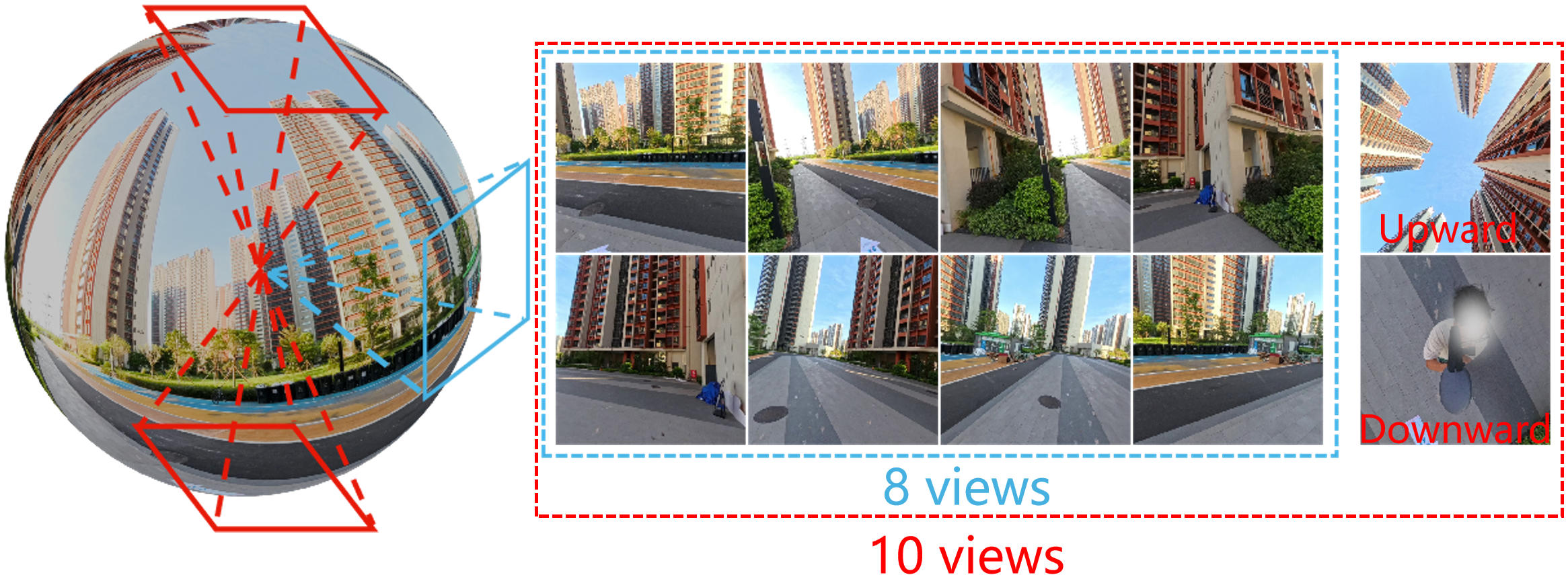}
    \caption{View decomposition strategies. The 8 views consists of uniformly spaced views along the horizontal direction, ensuring full horizontal coverage. The 10 views setup extends this by adding one upward and one downward view.}
    \label{fig:split}
\end{figure}

\begin{figure*}[!ht]
    \centering
    \includegraphics[width=1\linewidth]{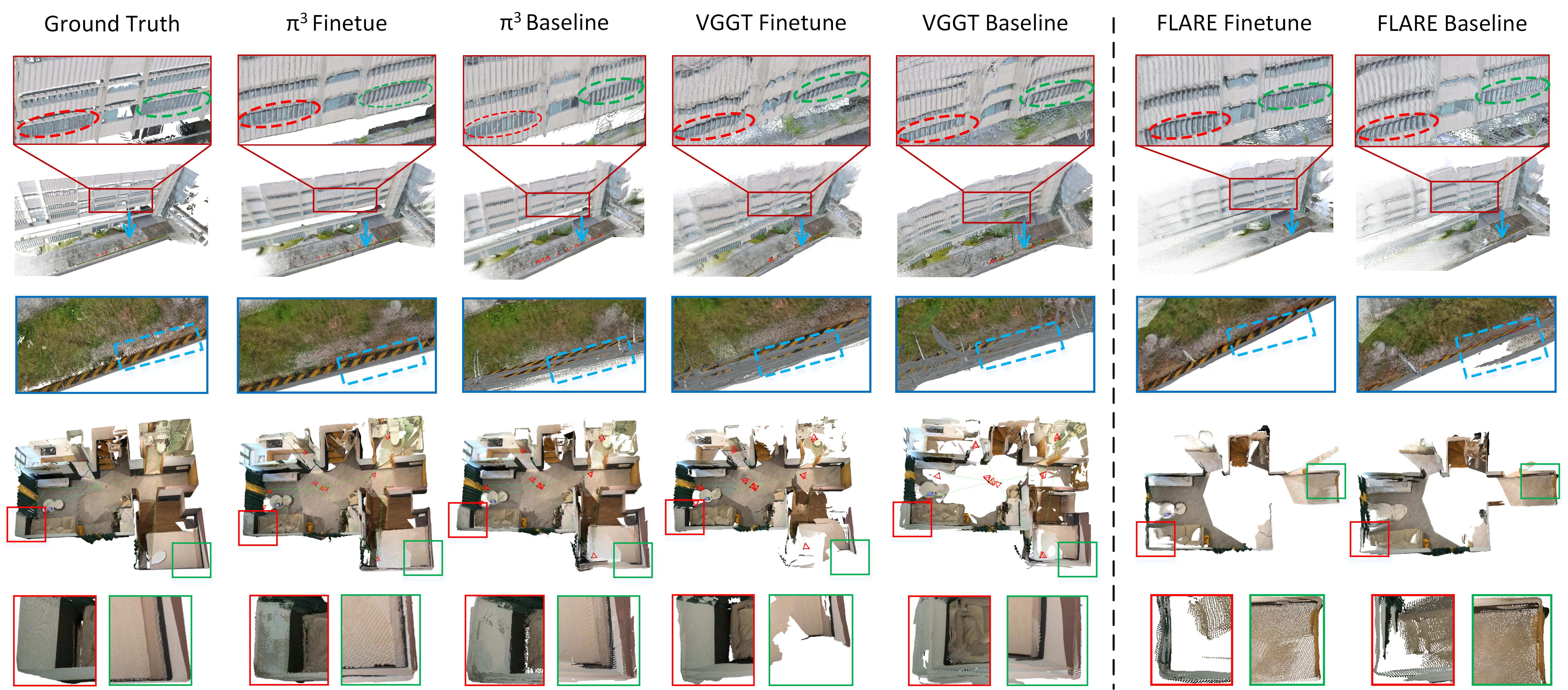}
    \caption{Visualization of baseline models fine-tuned on Holo360D. The blue arrows indicate viewpoints selected for zoom-in views.}
    \label{fig:algorithm_comparison}
\end{figure*}
\noindent\textbf{Depthmap Creation.} We produce both point and mesh depth maps by reprojecting the processed point cloud and mesh models into the 360° image space, respectively. For the mesh depth map, the viewing ray corresponding to each pixel is derived from the equirectangular projection. The depth at each pixel is defined as the Euclidean distance from the camera center to the nearest valid ray–mesh intersection. For the point depth map, the discrete point cloud fails to capture occlusions. The mesh depth maps from mesh are used to determine point cloud visibility. The procedure is as follows: the point cloud is projected to the panoramic image space, and the nearest point depth is retained per pixel to form a sparse depth map. This map is then compared with the dense depth map; points with greater depth values than their mesh-map counterparts are marked as occluded. The visible points are retained in the sparse depth map as the final depth output.

This three-step refinement yields accurate and complete point clouds, meshes, and depth maps.
\subsection{Datasets Statistics and Characteristics.}
\label{sec:Analysis of the Dataset}
To assess Holo360D against existing panoramic datasets, we report statistics of Holo360D from five perspectives: viewpoint sampling density, depth completeness, alignment error, point cloud reconstruction accuracy, and spatial coverage.

\noindent\textbf{Viewpoint Sampling Density.} Viewpoint sampling density is an important metric for evaluating the density and continuity of viewpoint distributions in multi-view 3D datasets, as dense and continuous viewpoint distributions can support more diverse 3D tasks. Holo360D provides the most continuous trajectory with an average sampling distance of $0.29\,\text{m}$, compared to $1.01\,\text{m}$ for KITTI-360, and $0.49\,\text{m}$ for 360Loc.

\noindent\textbf{Depth Completeness.} Depth completeness is defined as the proportion of valid depth pixels to the total number of pixels in each depth map. A higher completeness value indicates more complete and denser depth ground truth, which benefits the training and evaluation of 3D algorithms.

We assess the depth completeness of indoor and outdoor environments separately. For indoor scenes, we first compute the mean depth completeness for each scene, and then calculate the overall average across all scenes. For outdoor scenes, the calculation is similar, but sky regions are excluded from the calculation because depth of sky is invalid. To achieve this, we employ yoloe~\cite{wang2025yoloe} to obtain sky masks for all panoramic frames and compute the depth completeness only within the non-sky regions.

As reported in ~\cref{tab:summary_datasets}, Holo360D provides the highest depth completeness with $0.86$ indoors and $0.82$ outdoors, outperforming other datasets such as Stanford2D3D with $0.72$ indoors, 360Loc with 0.62 indoors and 0.7 outdoors.
The higher depth completeness stems from our dense viewpoint coverage and comprehensive post-processing pipeline. As shown in~\cref{fig:mp3d-holo360d--mesh-pose}, our mesh models exhibit improved geometric completeness, leading to more complete depth maps.

\noindent\textbf{Alignment Error.} Alignment error between panoramas and depth maps is essential for both algorithm training and evaluation. Following HELVIPAD~\cite{zayene2025helvipad}, we employ a manual point-selection strategy to assess alignment error. Specifically, we compute the pixel-wise Euclidean distance between 200 randomly selected depth points and their corresponding image points at visually salient locations. The average pixel error over these samples provides a quantitative measure of alignment error; all datasets are evaluated at a unified resolution. As reported in ~\cref{tab:summary_datasets}, Holo360D achieves superior alignment precision, with a mean error of 5.03 pixels, outperforming existing panoramic 3D datasets.


\noindent\textbf{Point Cloud Reconstruction Accuracy.} Point cloud reconstruction accuracy is an important metric for evaluating the geometric reliability of 3D dataset, as it directly determines the accuracy of depth maps. We evaluate the point cloud reconstruction accuracy by calculating the Root Mean Squared Error (RMSE) between the reconstructed point cloud and the ground truth. As illustrated in ~\cref{fig:pointcloud_reconstruction_accuracy}, to obtain ground truth, we use a laser rangefinder (with a $2\,\text{mm}$ measurement precision) to measure several key geometric dimensions in an indoor and an outdoor scene. The same dimensions are extracted from the 3D point clouds reconstructed by the scanner, and the RMSE between the two sets of data is reported as measurement errors. The point cloud reconstruction accuracy is \textbf{4.5\,mm} for indoor scenes and \textbf{7.0\,mm} for outdoor scenes.

\noindent\textbf{Spatial Coverage.} Holo360D also features long trajectories and broad spatial coverage, with up to 40,000 \text{m²} for a single scene and a maximum trajectory length of $5\,\text{km}$, supporting long-sequence panoramic 3D tasks. Across all scenes, Holo360D covers 190,000 m² of area and 31.5 km of trajectory, collected over 19 hours of on-site acquisition.



%% file: 4_experiment.tex
\section{Experiments} \label{sec:exp}
In this section, we first introduce the benchmark metrics and datasets used in our experiments. We then evaluate three fine-tuning configurations: input representations, view decomposition strategies, and supervision types. Finally, based on the optimal configuration, we conduct cross-model evaluations and cross-dataset comparisons to assess the effectiveness of Holo360D.

\subsection{Benchmark Metrics and Datasets}
\noindent \textbf{Benchmark Metrics.}
We evaluate models on camera pose estimation and point map estimation. \textbf{(i) Pose estimation}, we assess it using two categories of metrics: angular accuracy and distance error.
For angular accuracy, following \cite{wang2023posediffusion, wang2025vggt, wang2024dust3r, wang2025pi}, we compute the relative rotation accuracy (RRA) and relative translation accuracy (RTA) between consecutive frames. Furthermore, we compute the Area Under the Curve (AUC) of the $\min(\text{RRA}, \text{RTA})$–versus–threshold curve. Following \cite{wang2025continuous,zhang2024monst3r,zhao2022particlesfm}, we evaluate trajectory precision using Absolute Trajectory Error (ATE), Relative Pose Error-translation (RPE$_t$), and Relative Pose Error-rotation (RPE$_r$). \textbf{(ii) Point Map Estimation.} To evaluate the quality of multi-view point cloud reconstruction, we follow the protocol in \cite{wang2025continuous}. Predicted point maps are first coarsely aligned with ground truth using a similarity (Sim(3)) transformation computed via the Umeyama algorithm, followed by refinement using Iterative Closest Point (ICP) to ensure accurate alignment. After registration, we report two standard metrics: Accuracy (Acc.) and Completion (Comp.). These follow prior works \cite{azinovic2022neural, wang20243d, wang2025continuous, wang2024dust3r}.

\noindent \textbf{Datasets.} 
For our experimental studies, we  strictly divide Holo360D into training and test sets based on scene divisions to prevent potential data leakage, using 60 scenes for training and 15 scenes for testing. For each scene, we uniformly sample one-quarter of the data for training and testing in this experiment.
\begin{table*}[t]
\small
    \setlength\tabcolsep{1pt}
    \centering 
    \renewcommand\arraystretch{1.2}
    \caption{Comparison of fine-tuning performance under different view decomposition strategies and depth supervision types. \textcolor{red}{Red} and \textcolor{green}{green} denote the best and second-best results, respectively.}
     \resizebox{\textwidth}{!}{
\begin{tabular}{lcccccc|cc|cccc}
\toprule
\multirow{2}{*}{} 
  & \multicolumn{3}{c}{Indoor} 
  & \multicolumn{3}{c|}{Outdoor} 
  & \multicolumn{1}{c}{Indoor} 
  & \multicolumn{1}{c|}{Outdoor}
  & \multicolumn{2}{c}{Indoor} 
  & \multicolumn{2}{c}{Outdoor} \\
\cmidrule(lr){2-4} \cmidrule(lr){5-7} \cmidrule(lr){8-8} \cmidrule(lr){9-9} \cmidrule(lr){10-11} \cmidrule(lr){12-13}
& ~~ATE$\downarrow$~~ & ~~RPE$_t$$\downarrow$~~ & ~~RPE$_r$$\downarrow$~~ 
& ~~ATE$\downarrow$~~ & ~~RPE$_t$$\downarrow$~~ & ~~RPE$_r$$\downarrow$~~ 
& \scriptsize AUC@30$\uparrow$ 
& \scriptsize AUC@30$\uparrow$ 
& ~~Acc.$\downarrow$~~ & ~~Comp.$\downarrow$~~ 
& ~~Acc.$\downarrow$~~ & ~~Comp.$\downarrow$~~ \\
\midrule
$\pi^3$ (baseline) & 0.093 & 0.112 & 2.034 & 0.201 & 0.228 & 0.960 & 0.733 & 0.758 & 0.052 & 0.033 & 0.293 & 0.221 \\
~~~~+ ours, 8 views, mesh & \textcolor{red}{\textbf{0.014}} & \textcolor{red}{\textbf{0.014}} & \textcolor{red}{\textbf{0.295}} & \textcolor{red}{\textbf{0.065}} & \textcolor{red}{\textbf{0.049}} & \textcolor{red}{\textbf{0.208}} & \textcolor{red}{\textbf{0.837}} & \textcolor{red}{\textbf{0.818}} & \textcolor{red}{\textbf{0.024}} & \textcolor{red}{\textbf{0.018}} & \textcolor{green}{\textbf{0.194}} & \textcolor{green}{\textbf{0.152}} \\
~~~~+ ours, 10 views, mesh & 0.021 & 0.035 & 0.795 & 0.100 & 0.154 & 0.623 & 0.824 & 0.811 & 0.027 & 0.020 & 0.208 & 0.161 \\
~~~~+ ours, 8 views, point & 0.054 & 0.046 & 1.255 & 0.077 & 0.066 & 0.271 & 0.793 & 0.815 & 0.045 & 0.040 & 0.204 & 0.159 \\

~~~~+ ours, 8 views, mesh+point & \textcolor{green}{\textbf{0.015}} & \textcolor{red}{\textbf{0.014}} & \textcolor{green}{\textbf{0.304}} & \textcolor{green}{\textbf{0.071}} & \textcolor{green}{\textbf{0.054}} & \textcolor{green}{\textbf{0.225}} & \textcolor{green}{\textbf{0.834}} & \textcolor{green}{\textbf{0.816}} & \textcolor{red}{\textbf{0.024}} & \textcolor{red}{\textbf{0.018}} & \textcolor{red}{\textbf{0.189}} & \textcolor{red}{\textbf{0.146}} \\
\bottomrule
\end{tabular}}
\label{tab:ablation}
\end{table*}
\begin{table*}[t]
\small
    \setlength\tabcolsep{10pt}
    \centering 
    \renewcommand\arraystretch{1.2}
    \caption{Comparative experimental results across multiple models. Fine-tuning on Holo360D consistently improves performance over the corresponding baselines.}
     \resizebox{1.0\textwidth}{!}{
\begin{tabular}{lcccccc|cc|cccc}
\toprule
\multirow{2}{*}{} 
  & \multicolumn{3}{c}{Indoor} 
  & \multicolumn{3}{c|}{Outdoor} 
  & \multicolumn{1}{c}{Indoor} 
  & \multicolumn{1}{c|}{Outdoor}
  & \multicolumn{2}{c}{Indoor} 
  & \multicolumn{2}{c}{Outdoor} \\
\cmidrule(lr){2-4} \cmidrule(lr){5-7} \cmidrule(lr){8-8} \cmidrule(lr){9-9} \cmidrule(lr){10-11} \cmidrule(lr){12-13}
& ATE$\downarrow$ & RPE$_t$$\downarrow$ & RPE$_r$$\downarrow$ 
& ATE$\downarrow$ & RPE$_t$$\downarrow$ & RPE$_r$$\downarrow$ 
& \scriptsize AUC@30$\uparrow$ 
& \scriptsize AUC@30$\uparrow$ 
& Acc.$\downarrow$ & Comp.$\downarrow$ 
& Acc.$\downarrow$ & Comp.$\downarrow$ \\
\midrule
$\pi^3$ (baseline) & 0.093 & 0.112 & 2.034 & 0.201 & 0.228 & 0.960 & 0.733 & 0.758 & 0.052 & 0.033 & 0.293 & 0.221 \\
$\pi^3$ (finetune) & \textbf{0.014} & \textbf{0.014} & \textbf{0.295} & \textbf{0.065} & \textbf{0.049} & \textbf{0.208} & \textbf{0.837} & \textbf{0.818} & \textbf{0.024} & \textbf{0.018} & \textbf{0.194} & \textbf{0.152} \\
\midrule
VGGT (baseline) & 0.231 & 0.259 & 5.000 & 0.382 & 0.461 & 2.149 & 0.596 & 0.697 & 0.096 & 0.036 & 0.495 & 0.298 \\
VGGT (finetune) & \textbf{0.084} & \textbf{0.107} & \textbf{2.192} & \textbf{0.229} & \textbf{0.259} & \textbf{1.174} & \textbf{0.756} & \textbf{0.761} & \textbf{0.059} & \textbf{0.021} & \textbf{0.376} & \textbf{0.169} \\
\midrule
FLARE (baseline) & --- & --- & --- & --- & --- & --- & 0.256 & 0.345 & 0.133 & 0.111 & 0.852 & 0.848 \\
FLARE (finetune) & --- & --- & --- & --- & --- & --- & \textbf{0.515} & \textbf{0.479} & \textbf{0.031} & \textbf{0.026} & \textbf{0.365} & \textbf{0.263} \\

\bottomrule
\end{tabular}
}
\label{tab:algorithm}
\end{table*}
\subsection{Benchmark on Different Fine-Tuning Configurations} \label{sec:abl}

Since $\pi^3$ ~\cite{wang2025pi} outperforms other methods in zero-shot performance for panoramic multi-view 3D reconstruction, we conduct experiments to validate the fine-tuning strategies on this framework.

\noindent \textbf{Input Representations Evaluation.} 
We explore two input representations for fine-tuning feed-forward 3D reconstruction: (i) directly using panoramic images, and (ii) splitting panoramas into multiple perspective views. 
The experimental results, shown in Tab.~\ref{tab:Input type} and Fig.~\ref{fig:inputRepresenting}, indicate that the fine-tuning setting using split views produces more coherent and complete 3D structures. This is because feed-forward 3D models are designed for perspective images \cite{wang2025vggt,wang2025pi,wang2024dust3r,yang2025fast3r}, and split views mitigate the effects of spherical distortions. Panoramic input yields pose estimation results better than the baseline, but the point cloud reconstruction quality is suboptimal, even worse than the baseline. Therefore, we argue that training a perspective-based feed-forward model directly on panoramic data is suboptimal. Model adaptation is required to effectively handle spherical distortion, such as introducing panoramic rays to enhance geometric attention and designing a panoramic loss with latitude awareness to improve geometric supervision.

\noindent \textbf{View Decomposition Strategies Evaluation.} We evaluate two view decomposition strategies: (i) the 8 views configuration, which ensures full horizontal coverage, and (ii) the 10 views configuration, which adds upward and downward views for complete vertical coverage as shown in ~\cref{fig:split}. The comparison results, shown in Tab.~\ref{tab:ablation} and Fig.~\ref{fig:Ablatioion}, demonstrate that the 8 views outperforms the 10 views in pose and point cloud estimation.

The multiple views can complement each other’s missing field of view, allowing the 8 views configuration to maintain reconstruction integrity. In contrast, the 10 views setup introduces challenges due to dynamic operators in the downward view and low-texture regions, such as the ceiling, in the upward view, complicating consistent point cloud and pose estimation. Therefore, we conclude that the upward and downward views negatively affect handheld 360°  camera multi-view reconstruction. 
\begin{table*}[t]
\small
    \setlength\tabcolsep{2pt}
    \centering 
    \renewcommand\arraystretch{1.2}
    \caption{Comparison of $\pi^3$ finetuned on Holo360D vs. Matterport3D.}
     \resizebox{1.0\textwidth}{!}{
\begin{tabular}{lcccccc|cc|cccc}
\toprule
\multirow{2}{*}{} 
  & \multicolumn{3}{c}{Indoor} 
  & \multicolumn{3}{c|}{Outdoor} 
  & \multicolumn{1}{c}{Indoor} 
  & \multicolumn{1}{c|}{Outdoor}
  & \multicolumn{2}{c}{Indoor} 
  & \multicolumn{2}{c}{Outdoor} \\
\cmidrule(lr){2-4} \cmidrule(lr){5-7} \cmidrule(lr){8-8} \cmidrule(lr){9-9} \cmidrule(lr){10-11} \cmidrule(lr){12-13}
& ~~ATE$\downarrow$~~ & ~~RPE$_t$$\downarrow$~~ & ~~RPE$_r$$\downarrow$~~ 
& ~~ATE$\downarrow$~~ & ~~RPE$_t$$\downarrow$~~ & ~~RPE$_r$$\downarrow$~~ 
& \scriptsize AUC@30$\uparrow$ 
& \scriptsize AUC@30$\uparrow$ 
& ~~Acc.$\downarrow$~~ & ~~Comp.$\downarrow$~~ 
& ~~Acc.$\downarrow$~~ & ~~Comp.$\downarrow$~~ \\
\midrule
$\pi^3$ (baseline) & 0.093 & 0.112 & 2.034 & 0.201 & 0.228 & 0.960 & 0.733 & 0.758 & 0.052 & 0.033 & 0.293 & 0.221 \\
$\pi^3$ (Matterport3D) & 0.428 & 0.320 & 6.001 & 0.987 & 0.796 & 3.417 & 0.299 & 0.467 & 0.477 & 0.400 & 2.803 & 3.579\\
$\pi^3$ (Holo360D) & \textbf{0.014} & \textbf{0.014} & \textbf{0.295} & \textbf{0.065} & \textbf{0.049} & \textbf{0.208} & \textbf{0.837} & \textbf{0.818} & \textbf{0.024} & \textbf{0.018} & \textbf{0.194} & \textbf{0.152} \\

\bottomrule
\end{tabular}}
\label{tab:algorithm2}
\end{table*}
\begin{figure*}[t]
    \centering
    \includegraphics[width=0.8\linewidth]{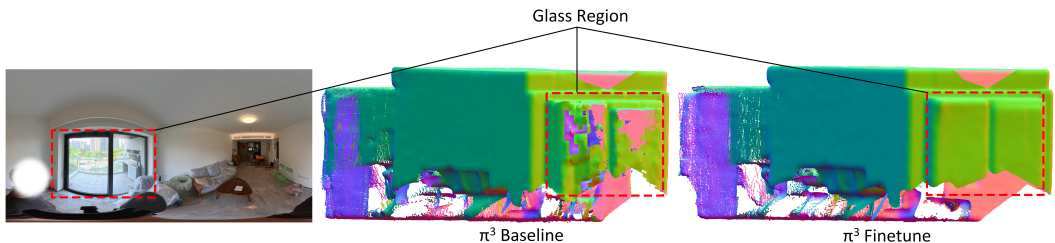}
    \caption{Comparison of reconstructions in glass regions before and after finetuning. The finetuned $\pi^3$ yields more complete and accurate geometry in transparent surfaces such as glass.}
    \label{fig:glass}
\end{figure*}

\begin{figure}[t]
    \centering
    \includegraphics[width=1.0\columnwidth]{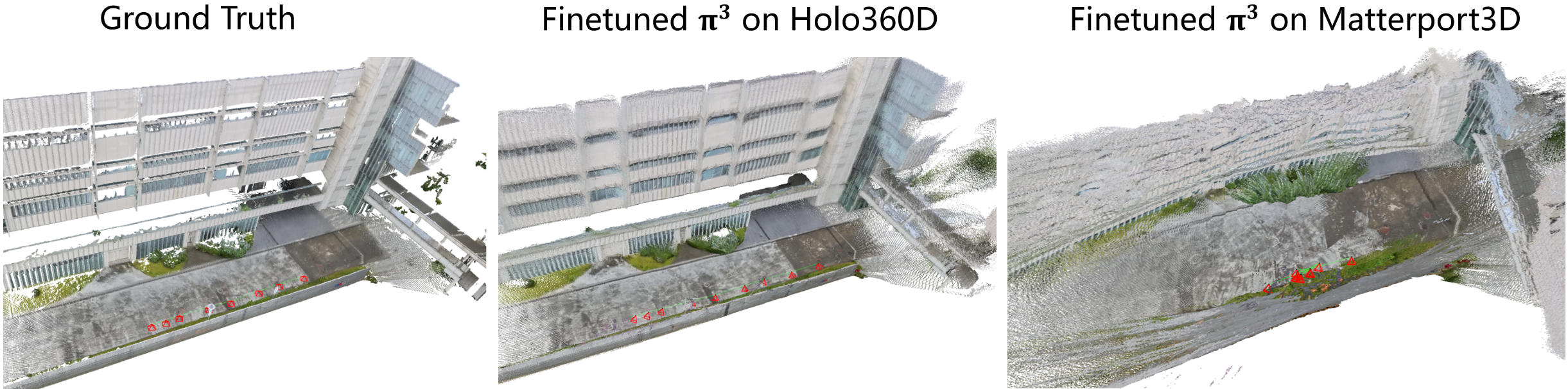}
    \caption{Finetuning $\pi^3$~on different datasets. Fine-tuning on Holo360D enables more accurate and complete reconstruction results than finetuning on Matterport3D.}
    \label{fig:Matterport3D_result}
\end{figure}

\noindent \textbf{Depth Supervision Types Evaluation.} We evaluate three depth supervision configurations: (i) training with mesh depth, (ii) training with point depth, and (iii) training with mesh depth followed by fine-tuning with point depth.

As illustrated in Fig.~\ref{fig:Ablatioion} and Tab.~\ref{tab:ablation}, our experiments show that mesh depth supervision achieves the highest pose estimation accuracy and second-highest point cloud accuracy. This is due to both the strong geometric supervision it provides and the increased 3D-to-2D correspondences it establishes. Fine-tuning with point depth following training in mesh depth further improves geometric accuracy, particularly in complex structures like railings in \cref{fig:Ablatioion} row 2. However, reduced geometric supervision lowers pose estimation accuracy.
Overall, mesh depth provides strong continuous geometric constraints, point depth offers a more accurate evaluation of the model's performance, serving as an accurate ground truth.

Based on the evaluation of all configurations, we determine that the optimal setup is the 8 views setting combined with mesh depth for the next section's evaluation.

\subsection{Cross-Model Evaluation and Cross-Dataset Comparison} \label{sec:bench}
\noindent \textbf{Cross-Model Evaluation on Holo360D.}
Based on the conclusions drawn in Sec.~\ref{sec:abl}, we adopt mesh depth as supervision and use the 8 views configuration as input. We then compare three feed-forward 3D reconstruction models: $\pi^3$, VGGT, and FLARE. This evaluation aims to assess the cross-model generalization capability of our dataset. For VGGT, we follow the exact training and evaluation setup used for $\pi^3$. For FLARE, due to its high VRAM requirements, we limit each training and testing iteration to two panoramic frames. As EVO cannot align trajectories with the ground truth when only two pose points are available, the distance-based pose metrics are not reported for FLARE.

As shown in Tab.~\ref{tab:algorithm} and Fig.~\ref{fig:algorithm_comparison}, all models exhibit notable improvements over their baselines after finetuning on Holo360D, both in quantitative metrics and in qualitative performance. These consistent gains demonstrate that our dataset effectively enhances performance in panoramic 3D reconstruction across diverse model architectures.
Additionally, as shown 

\noindent in Fig.~\ref{fig:glass}, the finetuned $\pi^3$ model achieves better reconstruction quality on challenging transparent surfaces, such as glass windows. For a more intuitive comparison, we visualize the surface normals of the reconstructed point clouds. This improvement is attributed to our mesh hole filling process, which enables the model to receive effective supervision even in glass regions.

\noindent \textbf{Cross-Dataset Comparison.}
We select Matterport3D~\cite{chang2017matterport3d} as a representative of existing panoramic 3D datasets for comparison with our dataset. It provides the most accurate depth maps among existing panoramic 3D datasets and its re-rendered version also offers continuous three-view sequences, making it an ideal reference for comparison. As shown in Fig.~\ref{fig:Matterport3D_result} and \cref{tab:algorithm2}, the model fine-tuned on Holo360D significantly outperforms the one fine-tuned on Matterport3D, further highlighting the advantages of our dataset for this task.

%% file: 5_conclusions.tex
\section{Conclusions}
\label{sec:conclusion}
In this work, we introduced Holo360D, the pioneering large-scale real-world panoramic dataset characterized by its precisely aligned high-completeness depth maps and continuous trajectories. Our extensive benchmarking across various fine-tuning regimes yielded three pivotal insights: (i) training feed-forward 3D models on panoramic images is challenging; (ii) vertical perspectives (nadir and zenith) often introduce noise that detracts from multi-view reconstruction accuracy; and (iii) mesh-based depth serves as a superior supervisory signal compared to sparse point depth. The consistent performance leap observed across multiple feed-forward 3D models underscores the robust generalization and utility of Holo360D. These findings emphasize the imperative for panoramic-specific adaptations in feed-forward 3D reconstruction. We anticipate that Holo360D will serve as a cornerstone for the development and validation of next-generation feed-forward panoramic models.

%% file: X_suppl.tex
\clearpage

\twocolumn[{%
  \vspace{-1.1cm}
  \centering
  {\LARGE\bfseries Supplementary Material}
  \vspace{0.5cm}
}]
\setcounter{page}{1}

\section{Equipment Details} \label{sec:equipmentdetails}
The data acquisition device integrates a LiDAR, RTK-GNSS, IMU, three pinhole cameras, and a 360° camera. The LiDAR offers a 360° × 270° (Horizontal × Vertical) field of view, with a sensing range from 0.05 m to 120 m. It captures point clouds at 320,000 points per second, achieving an absolute precision of 5 cm and a relative precision of 1 cm. The IMU operates at 200 Hz. The 360° camera records video at 5760 × 2880 resolution and 24 fps, using a 1/2 inch image sensor to produce high-resolution panoramic images. The specifications of the device are summarized in Tab. ~\ref{tab:scanner_params}.

\begin{table}[h]
\centering
\caption{Specific Parameters of the 3D Scanner.}
\label{tab:scanner_params}
\scriptsize
\setlength\tabcolsep{4pt}
\renewcommand{\arraystretch}{1.2}
\begin{tabular}{@{}lcc@{}}
\toprule
\textbf{Module} & \textbf{Key Parameters} & \textbf{Value} \\
\midrule
\multirow{6}{*}{\textbf{LiDAR}} 
  & Absolute accuracy & 5\,cm \\
  & Relative accuracy & 1\,cm \\
  & Field of view & $360^\circ \times 270^\circ$ \\
  & Measurement range & $0.05\text{--}120$\,m \\
  & Point frequency & $320{,}000$\,points/s \\
  & Number of laser channels & 16 \\
\midrule
\multirow{2}{*}{\textbf{RTK-GNSS}} 
  & Horizontal accuracy & $\pm(8 + 1\times 10^{-6} D)$\,mm \\
  & Elevation accuracy & $\pm(15 + 1\times 10^{-6} D)$\,mm \\
\midrule
\textbf{IMU} 
  & IMU frequency & $200$\,Hz \\
\midrule
\multirow{2}{*}{\textbf{360$^\circ$ Camera}} 
  & Video resolution & $5760\times 2880@24$\,fps \\
  & Sensor size & $1/2$ inch \\
\bottomrule
\end{tabular}
\vspace{-0.5cm}
\end{table}

\section{Dataset Details}
\label{sec:datasetdetails}
\subsection{Challenging Scenes} \label{sec:challengingScenes}
\begin{figure*}[p]
    \centering
    \includegraphics[width=0.9\linewidth]{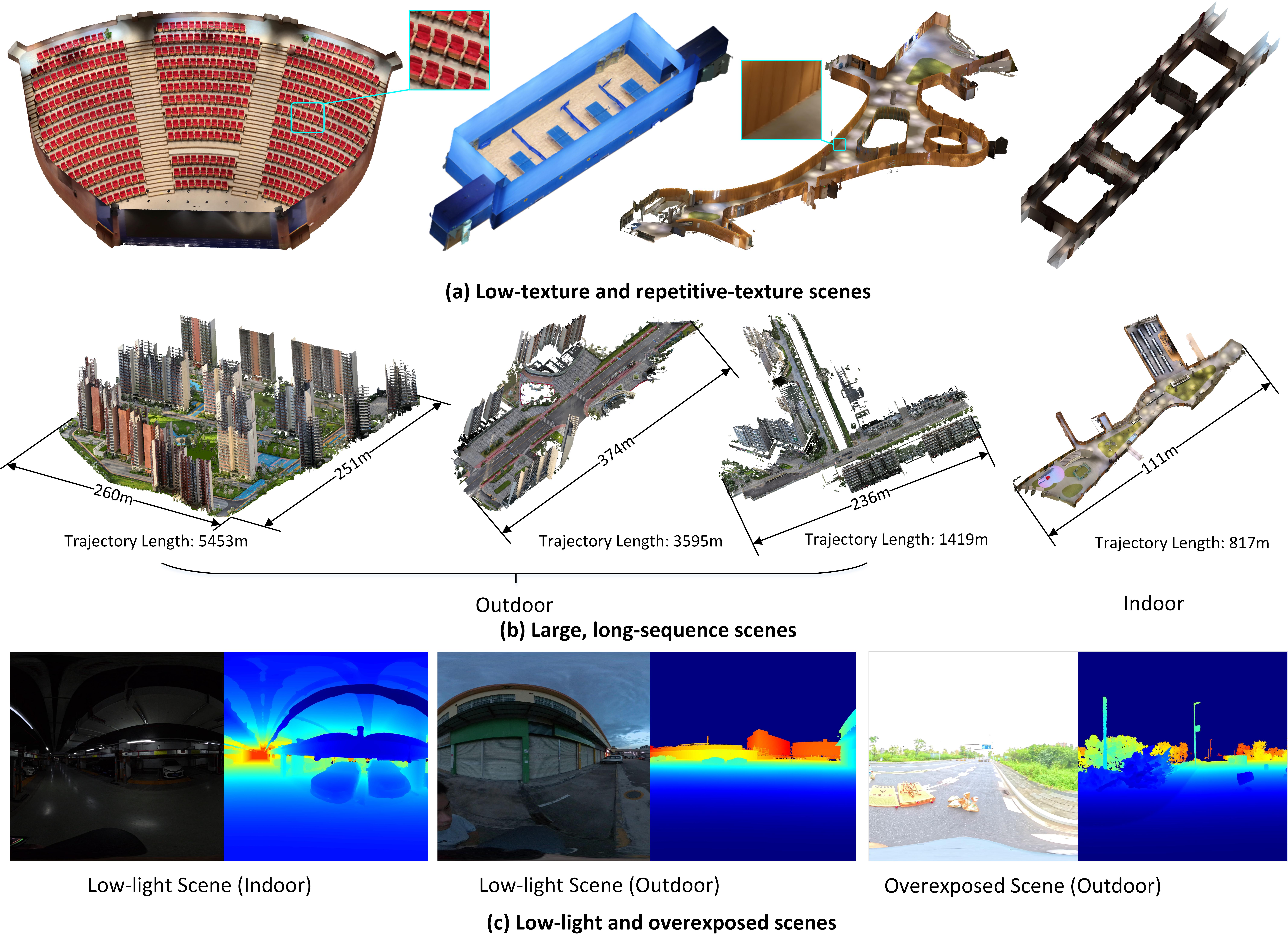}
    \caption{Challenging scenes. Our dataset includes (a) low-texture and repetitive-texture scenes, (b) large, long-sequence scenes, and (c) low-light and overexposed scenes, providing a comprehensive basis for testing the performance of panoramic 3D reconstruction algorithms.}
    \label{fig:challengingScenes}
\end{figure*}

As shown in Fig.~\ref{fig:challengingScenes}, our dataset includes several challenging scenes, including (a) low-texture and repetitive-texture scenes, (b) large, long-sequence scenes, as well as (c) low-light and overexposed scenes. These challenging environments provide a robust basis for thoroughly evaluating the performance of panoramic 3D reconstruction algorithms.

\subsection{Point Cloud Visualization from Depth Maps} \label{sec:point cloud visualization}
\begin{figure*}[p]
    \centering
    \includegraphics[width=0.9\linewidth]{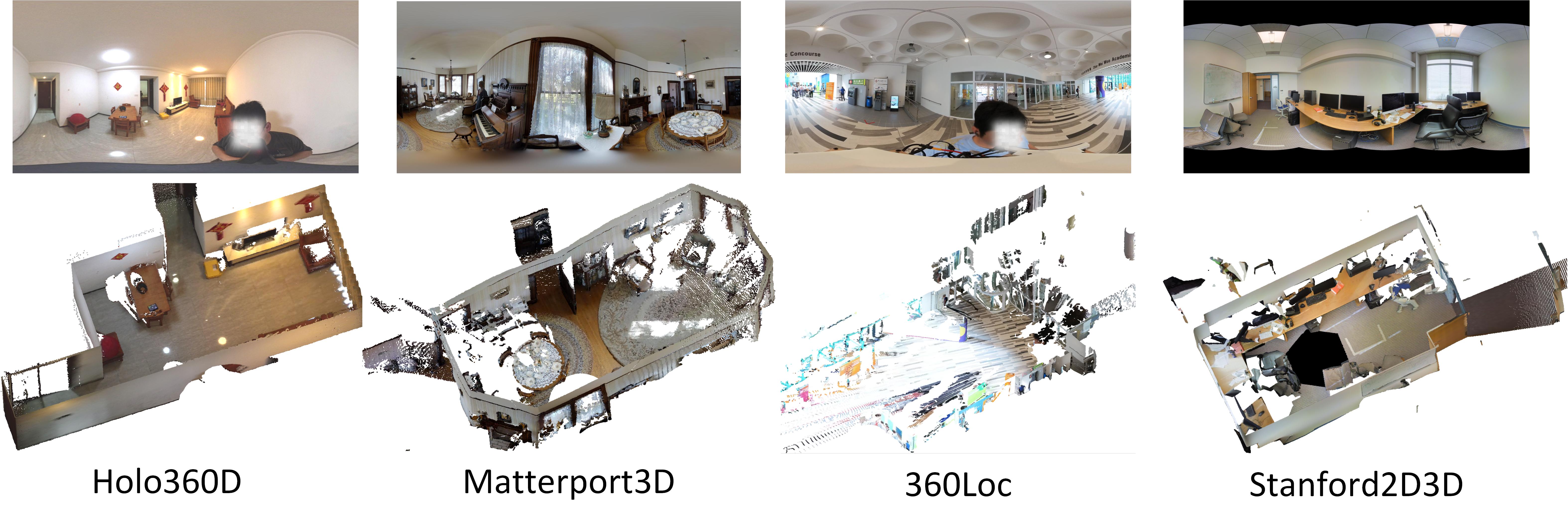}
    \caption{Comparison of single-frame point clouds.}
    \label{fig:singleviewpoint}
\end{figure*}

To further compare the quality of depth maps provided by Holo360D with existing datasets, we project single-frame depth maps into point clouds. As shown in \cref{fig:singleviewpoint}, the point clouds of Holo360D exhibit higher fidelity and completeness.

\section{Experiment  Settings} \label{sec:expdetail}

During training, we adopt a dynamic batch size following $\pi^3$. We sample \(n\) panoramic images (\(n \in [3,6]\)) from a randomly selected window of a sequence and decompose each panorama into eight perspective views. Thus, each training batch contains 24–48 perspective images, with at most 48 images processed on each GPU. We train each model using 4 NVIDIA A800 GPUs for 50 epochs, with each epoch consisting of 1,000 iterations.

\begin{figure*}[p]
    \centering
    \includegraphics[width=0.8\linewidth]{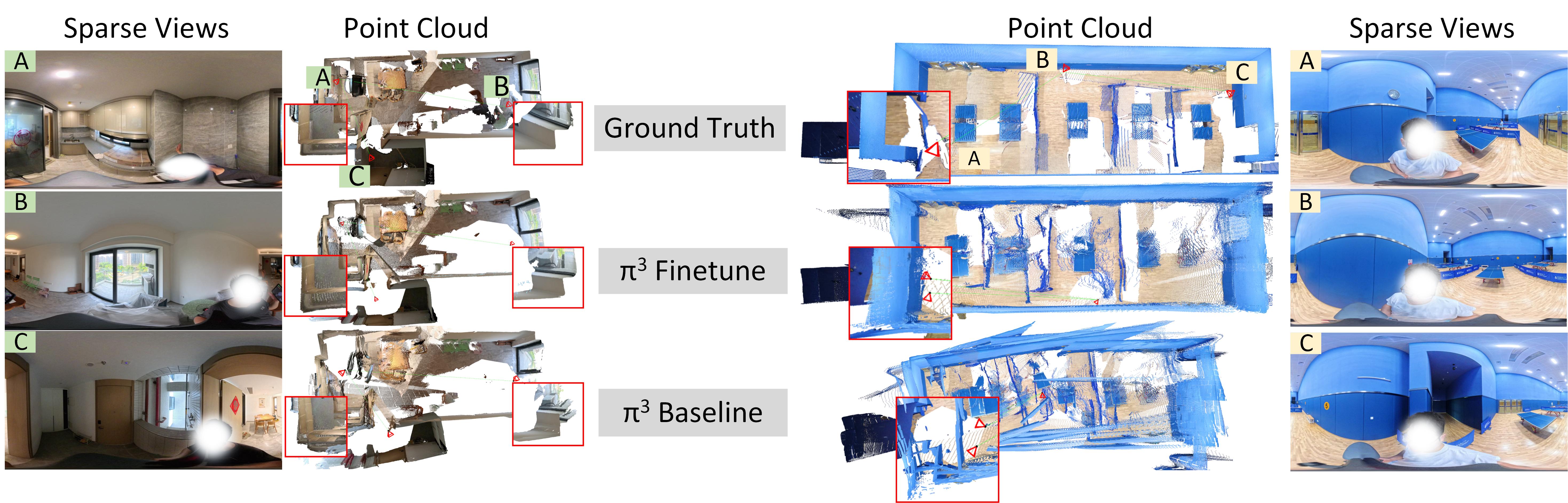}
    \caption{Qualitative comparison of sparse-view panoramic 3D reconstruction results. After finetuning with our dataset, the $\pi^3$ produces more consistent point clouds compared to the baselines. }
    \label{fig:sparse}
\end{figure*}

\begin{figure*}[p]
    \centering
    \includegraphics[width=0.7\linewidth]{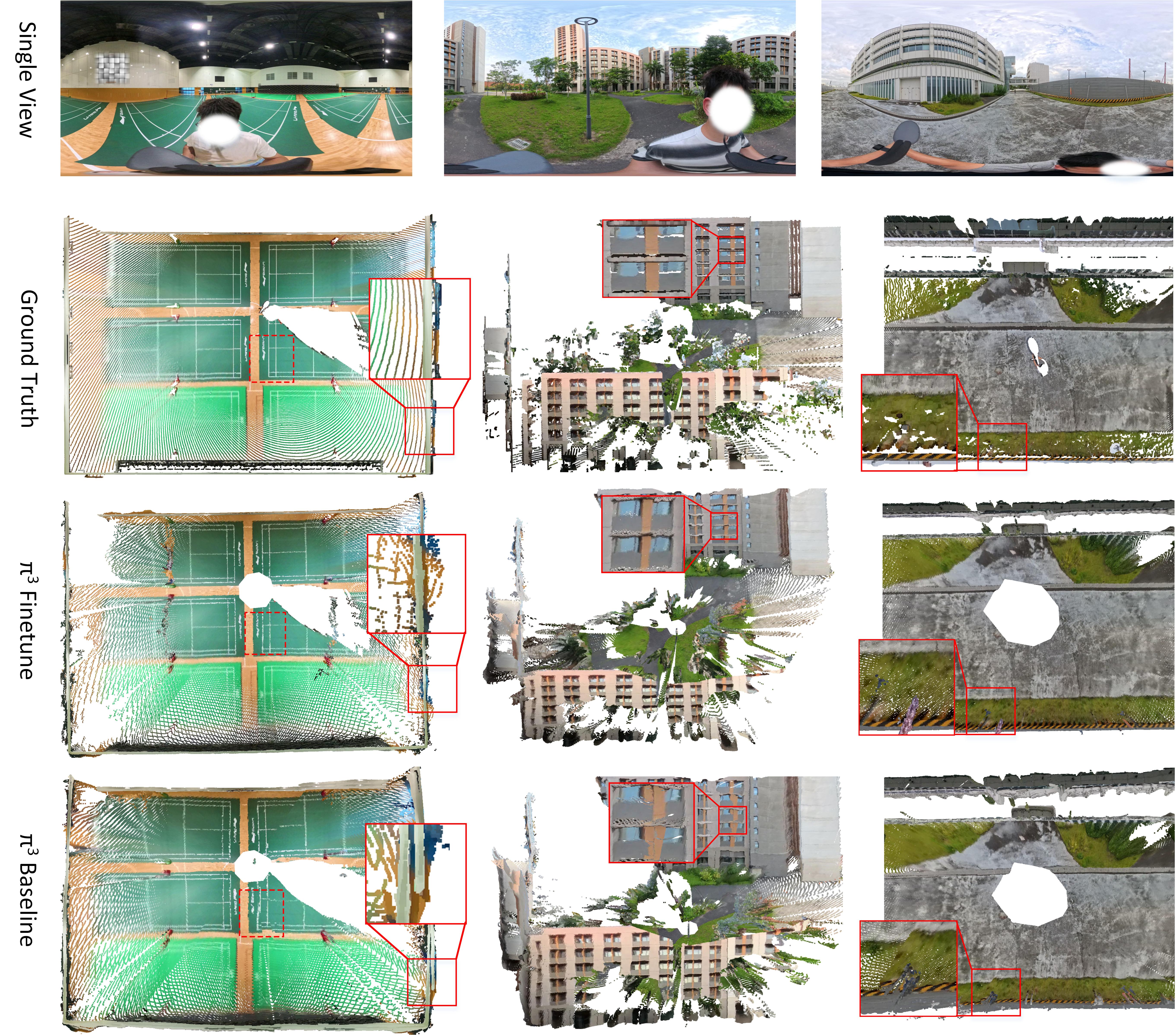}
    \caption{Qualitative comparison of single-view panoramic 3D reconstruction results. After finetuning with our dataset, the model produces more accurate point clouds compared to the baselines.}
    \label{fig:singleview}
\end{figure*}

\section{More Results}
\label{sec:more_results}

As discussed in Sec.~\ref{sec:bench} of the main paper, all models show improved quantitative and qualitative performance after finetuning on our dataset. To complement these findings, we further assess the qualitative performance of the finetuned $\pi^3$ model under diverse evaluation settings, including sparse-view and single-view panoramic reconstruction. 
As shown in Fig.~\ref{fig:sparse}, to perform sparse-view reconstruction, we selected two scenes from the test set and, for each scene, chose three sparse panoramic views labeled A, B, and C. The finetuned $\pi^3$ model demonstrates better pose accuracy and reconstruction quality. 

\begin{figure*}[p]
    \centering
    \includegraphics[width=0.7\linewidth]{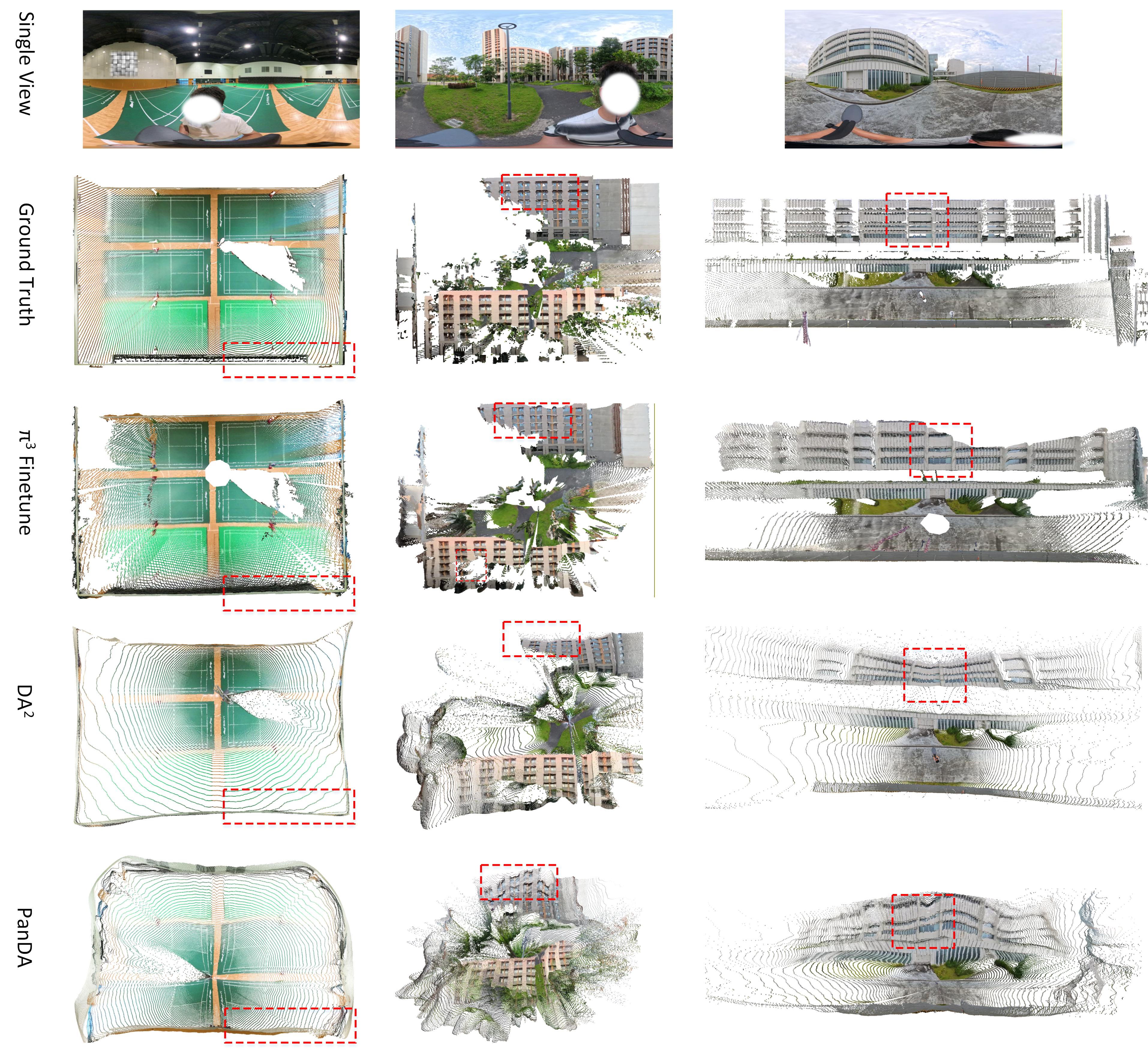}
    \caption{Comparison between the advanced panoramic monocular depth estimation models (DA$^2$ [1] and PanDA [2]) and the finetuned 3D reconstruction model ($\pi^3$). The finetuned $\pi^3$ achieves better geometric consistency across the reconstructed scenes.}
    \label{fig:360depth}
\end{figure*}

Meanwhile, we compare the reconstruction quality before and after finetuning in the single-view setting. As shown in Fig.~\ref{fig:singleview}, finetuning improves point cloud accuracy and significantly reduces layering artifacts. 
We further compare the quality of point clouds generated from single-frame panoramic images using the 3D reconstruction model (finetuned $\pi^3$) and the monocular depth estimation models (DA$^2$ [1] and PanDA [2]).
As shown in Fig.~\ref{fig:360depth}, we observe that the finetuned $\pi^3$ approach achieves higher geometric accuracy than other depth estimation methods. We attribute this improvement to two main factors. First, our dataset provides outdoor and long-range indoor scenes, enabling finetuned $\pi^3$ to generalize better to such environments. Second, although monocular depth estimation methods can effectively predict depth, distortions remain after converting depth maps 

\noindent into point clouds, leading to lower geometric fidelity compared with the 3D reconstruction results.

\section{Limitations and Future Work} \label{sec:limitations}
\begin{figure*}[p]
    \centering
    \includegraphics[width=0.9\linewidth]{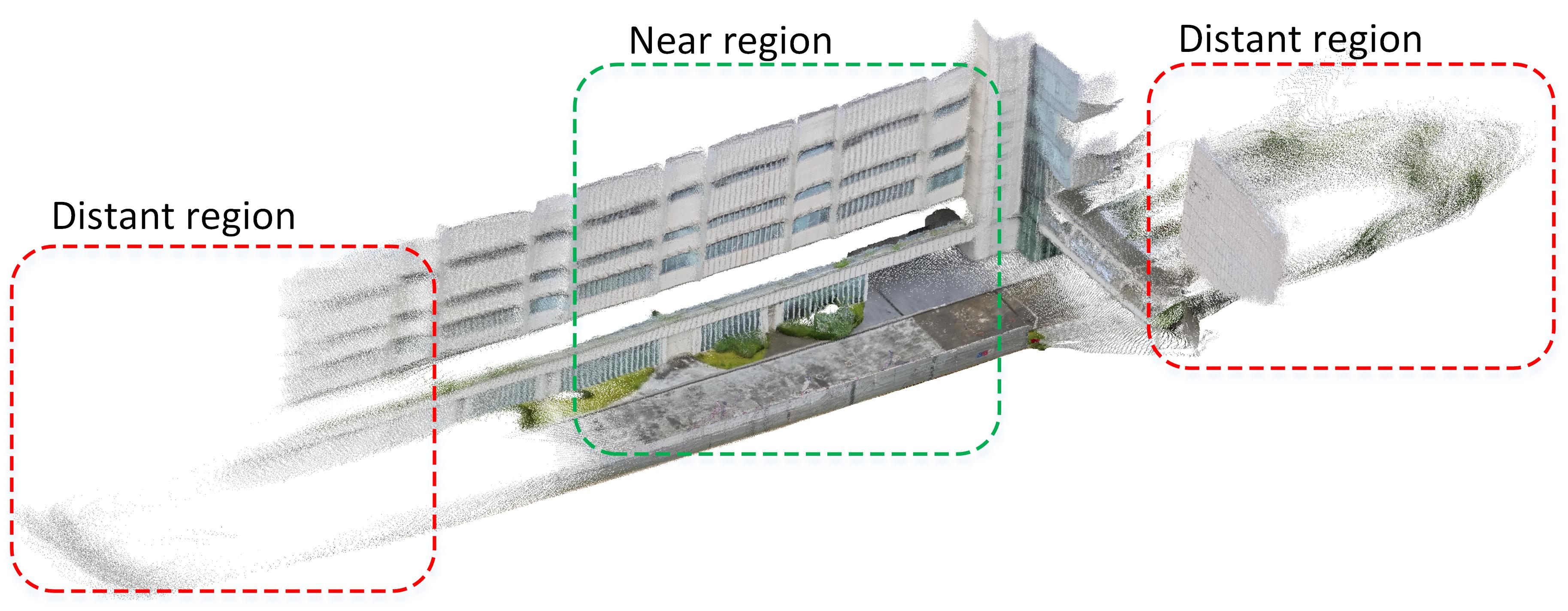}
    \caption{Degradation of reconstruction quality in distant regions.}
    \label{fig:distantRegion}
\end{figure*}

\noindent\textbf{Limitations.}
Although our dataset surpasses existing ones in both scale and quality and significantly improves the performance of fine-tuned models, one limitation remains: reconstruction quality degrades in distant regions. As shown in Fig.~\ref{fig:distantRegion}, while the finetuned model performs well in near regions, the geometric accuracy degrades noticeably in distant areas. This degradation is primarily due to insufficient supervision and reduced spatial resolution for long-range regions. Although the dataset maintains an approximately 1:1 balance between indoor and outdoor scenes, distant regions still cover only a small fraction of image pixels, even in outdoor settings. As a result, these regions receive less effective supervision during training.

\noindent\textbf{Future Work.}
To address the issues above, we plan to further expand the dataset with a particular focus on increasing the proportion of distant-region pixels. This will help enhance the model’s ability to learn long-range geometry and improve its generalization to diverse real-world scenarios.

